\documentclass[letterpaper, 10 pt, conference]{ieeeconf}  

\IEEEoverridecommandlockouts                              

\usepackage{graphicx}
\usepackage{epsfig} 
\usepackage{mathptmx} 
\usepackage{times} 
\usepackage{amsmath} 
\usepackage{amssymb}  
\usepackage{subcaption}
\usepackage{mathtools}
\usepackage{subcaption}
\usepackage{caption}
\usepackage{hyperref}
\usepackage{balance}

\usepackage{siunitx}
\usepackage{color}

\usepackage{balance}

\title{\LARGE \bf
Feedback Control for Autonomous Riding of Hovershoes \\ by a Cassie Bipedal Robot
} 

\author{Shuxiao Chen*, Jonathan Rogers*, Bike Zhang*, and Koushil Sreenath
\thanks{* These authors contributed equally to this work.}
\thanks{This work was supported by National Science Foundation Grant IIS-1834557 and Berkeley Deep Drive.}
\thanks{S.\ Chen, J.\ Rogers, B.\ Zhang and K.\ Sreenath are with the Department of Mechanical Engineering, University of California, Berkeley, CA, 94720, USA,
        {\tt\small \{shuxiao.chen, jr1397, bikezhang, koushils\}@berkeley.edu}}%
}

\begin{document}
\bstctlcite{IEEEexample:BSTcontrol}

\maketitle
\thispagestyle{empty}
\pagestyle{empty}

\begin{abstract}
Motivated towards achieving multi-modal locomotion, in this paper, we develop a framework for a bipedal robot to dynamically ride a pair of Hovershoes over various terrain. Our developed control strategy enables the Cassie bipedal robot to interact with the Hovershoes to balance, regulate forward and rotational velocities, achieve fast turns, and move over flat terrain, slopes, stairs, and rough outdoor terrain. Our sensor suite comprising of tracking and depth cameras for visual SLAM as well as our Dijkstra-based global planner and timed elastic band-based local planning framework enables us to achieve autonomous riding on the Hovershoes while navigating an obstacle course. We present numerical and experimental validations of our work.
\end{abstract}

\vspace{2mm}


\section{Introduction}
While locomotion using legs is efficient when traveling over rough and discrete terrain, wheeled locomotion is more efficient when traveling over flat continuous terrain \cite{c1}. Humans are able to optimize locomotion efficiency by using multiple locomotion modalities that comprise of not only being able to walk and run, but also being able to ride various micro-mobility platforms, such as Segways and Hovershoes. Enabling legged robots to autonomously ride on various personal mobility platforms will offer multi-modal locomotion capabilities, improving the efficiency of locomotion over various terrains. 

\begin{figure}[t]
    \centering
    \includegraphics[width=1.0\linewidth]{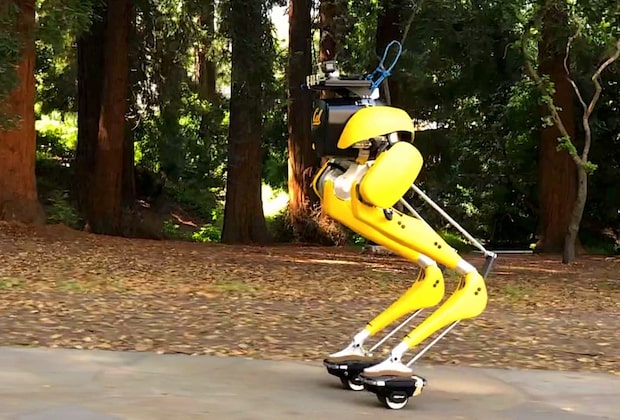}
    \caption{Snapshot of an experiment of the bipedal robot Cassie autonomously riding on Hovershoes.
    Experimental videos are at \href{https://youtu.be/b2fKBb_0iTo}{https://youtu.be/b2fKBb\_0iTo}. }
    \label{whole_system}
    \vspace{-5mm}
\end{figure}

Autonomous robots with multi-modal locomotion capabilities can have a big impact in the real-world from package delivery to security and surveillance to search and rescue missions \cite{c2}. In order to address this problem, we use the Cassie bipedal robot developed by Agility Robotics to autonomously ride Hovershoes, see Fig.~\ref{whole_system}.

\subsection{Challenges}
There are multiple challenges in developing an autonomous framework for Cassie to ride Hovershoes. First, Cassie is a complex, underactuated robot with $20$ degrees-of-freedom (DOFs), making it a difficult system to control. Second, the Hovershoes are a sensitive and highly dynamic platform that serve as two decoupled moving platforms under each of Cassie's feet and are hard to coordinate. Third, the algorithms we can develop neither have access to the internal states of the Hovershoes nor can they directly specify the torque inputs on the Hovershoes, since Cassie can only indirectly interact with the Hovershoes through the contact forces. Fourth, the Cassie-Hovershoes system needs to be robust to kinematic and dynamic variances arising from (i) manual initialization of the Hovershoes that result in an initial relative translation and orientation between the Hovershoes; (ii) manual placement of Cassie's feet on the Hovershoes resulting in off-center foot placement; and (iii) differences in the dynamic characteristics between the Hovershoes due to manufacturing. Fifth, the Cassie-Hovershoes system needs to be able to autonomously avoid obstacles, resulting in fast perception and planning demands.

We solve these challenges by designing controllers that assume the dynamics are decoupled along various degrees-of-freedom, strategically choosing a minimalist sensor suite for state estimation, and assuming kinematics is sufficient for path planning.  This motivates the simplest approach to solving the problem - using PD controllers on decoupled dynamics, using a VIO and depth sensor for state estimation and object detection, and simple global and local kinematic planners.  In the future, we will look at more complex controllers, estimators and planners.


\subsection{Related Work}

\subsubsection{Multi-Modal Locomotion}

There are some examples of multi-modal locomotion. For instance, DRC-HUBO+ is a bipedal robot that can have wheeled attachments at the knee joint of each leg \cite{c3}, enabling both legged and wheeled locomotion. Boston Dynamics' Handle robot\textemdash designed for box handling in warehouses\textemdash is another legged robot with wheels for feet \cite{c4}. ANYmal is yet other wheeled-legged robot used for traveling on irregular terrain \cite{c6}. While these robots can offer multi-modal locomotion, the integrated leg-wheel design is neither optimized for legged nor for wheeled mobility, resulting in complex and heavy feet.

Micro-mobility platforms\textemdash such as Segways and Hoverboards\textemdash allow legged robots without wheeled feet to also have multi-modal locomotion capabilities. This is advantageous since it keeps the legs lighter for faster motions during legged movement, while also having the option of wheeled mobility by riding on these platforms when needed.  
Segways and Hoverboards are relatively easy to ride since the platforms for either foot are connected for increased stability and turning is accomplished through simply leaning on a bar or differentially actuating the two connected platforms. However, these micro-mobility systems have limited abilities since they have a single platform for the user's feet, as discussed next.

\begin{figure}[]
    \centering
    \includegraphics[width=1.0\columnwidth]{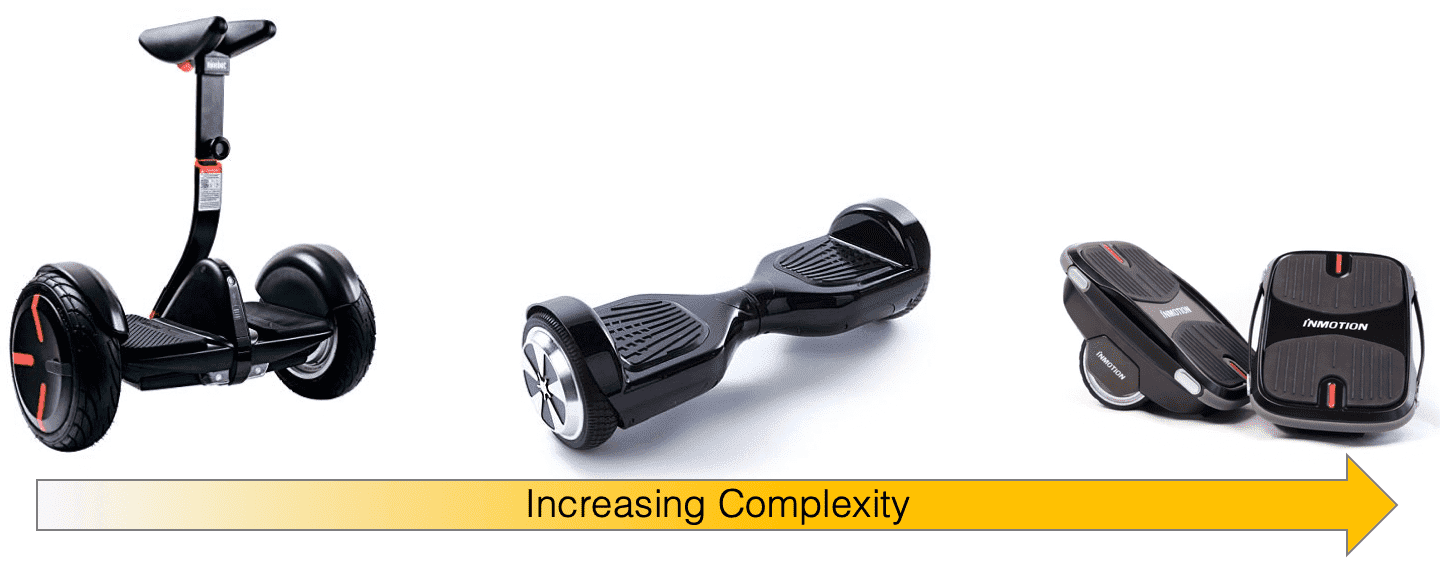}
    \caption{Segway, Hoverboard, and Hovershoes are shown from left to right. The complexity in terms of user control increases along the arrow head direction.}
    \label{segway}
    \vspace{-5mm}
\end{figure}

Hovershoes offer more versatile movements since each of the user's feet can move independently.  However, Hovershoes increase the riding complexity as they are significantly more sensitive and have no internal controller for turning. Though the Hovershoes are the most complex to ride, they offer the greatest versatility as each foot can move independently to avoid obstacles and go on uneven terrain, which is why we selected this platform for our research. 
We illustrate the degree of complexity of riding these products in Fig.~\ref{segway}.

\subsubsection{Controls}

There has been research on developing controls for balancing on unstable environments such as seesaws and Bongo Boards with robust lateral stabilization \cite{c7}. However, these environments do not translate; but rather, they only tilt.   
There has been recent work on balancing on platforms that both tilts and translates, a Hoverboard, through sequential online learning control \cite{c8}. Additionally, a balancing controller for Cassie worked in riding a Segway by regulating the center-of-mass (COM) position \cite{c9}. However, these control strategies, though, are only for rigidly connected wheeled platforms and do not work on Hovershoes, which are not connected and add additional degrees-of-freedom for more versatility. Our research aims to develop a control algorithm for a bipedal robot to balance on a highly dynamic and decoupled platform under each foot.

\subsubsection{Planning}

Generic planning algorithms are primarily categorized into sampling-based and search-based methods.
Sampling-based planning methods are computationally economical and offer high update rates.  
Nevertheless, feasible trajectories are not always guaranteed and there is some stochasticity in the choice of the plan with the same environmental layout potentially resulting in different plans \cite{c10}.
In contrast, search-based methods produce deterministic solutions, 
however, they may incur redundant computation due to their iterative searches \cite{c11}.

\subsubsection{Vision}

An accurate odometry source is of vital importance to robot movements and velocity tracking.  As a benchmark study reveals, VINS-Mono, a monocular VIO algorithm, with loop-closure detection offers decent accuracy and robustness and so does OKVIS, a stereo algorithm \cite{c12, c13}. 
Besides, there are also mature back-end mapping solutions to collaborate with the front-end VIO, e.g. ORB-SLAM2 which introduces a lightweight framework and RTAB-Map that uses OctoMap to achieve good memory management \cite{c15, c16}. 

For walking tasks on legged robots, there has been research on utilizing the contact between the feet and the static environment to optimize for odometry \cite{c18}. However, since in our case the feet are always in contact with the Hovershoes and the motion is relatively smooth, some standard robust VIO and SLAM solutions are sufficient for our scenario \cite{c19}.

\subsection{Contribution}
The contributions of our work thus are:

\subsubsection{Dynamics}
We developed a dynamical model of the integrated Cassie-Hovershoes system for numerical validation.

\subsubsection{Controls}
We designed a control strategy for bipedal robots to robustly balance on independently mobile and decoupled wheeled platforms. 

\subsubsection{Autonomy}
We developed a framework comprising of a vision system for odometry and obstacle detection, a path planner for online trajectory generation, and a control strategy for the Cassie bipedal robot to autonomously ride the Hovershoes and avoid obstacles. 

\subsubsection{Real-world Experiments}
We demonstrated the robustness of our controller for Cassie to robustly ride on Hovershoes subject to external perturbations on real-world terrain (flat and rough ground as well as stairs), track commanded translational and rotational velocities, perform turning maneuvers, and ride the Hovershoes in a wave pattern. 

\subsubsection{Simplicity}
We presented a solution that uses PD controllers designed assuming decoupled dynamics, an almost minimal sensor-suite for pose estimation and object detection, and kinematic planners to successfully solve the complex problem of making a bipedal robot autonomously ride a pair of Hovershoes.

\subsection{Organization}
The rest of the paper is organized as follows. Section \ref{sec:Dynamics} presents the dynamical model of the Cassie-Hovershoes system. Section \ref{sec:Control} describes our proposed controller design. Section \ref{sec:Simulation} presents the simulation results. The perception and planning are described in Section \ref{sec:Experimental_Setup}. Section \ref{sec:Experimental_Results} demonstrates experimental results with discussion.  Section \ref{sec:Limitations} discusses shortcomings of our work. Finally, Section \ref{sec:Conclusion} summarizes the work and provides thoughts on future work. 

\section{Dynamical Model of Cassie on Hovershoes} 
\label{sec:Dynamics}

Having established the need for multi-modal locomotion, described the challenges, and outlined our solution, we now present a dynamical model of Cassie and Hovershoes that also captures the interaction between each other.

\subsection{Dynamical Model of Cassie}

Cassie is a highly dynamic, under-actuated bipedal robot. Cassie has twenty DOFs as listed in \eqref{state_vector}:
\begin{equation}
\begin{split} 
q = [q_x, \ q_y, \ q_z, \ q_{\textrm{yaw}}, \ q_{\textrm{pitch}}, \ q_{\textrm{roll}}, \\
     q_{1L}, \ q_{2L}, \ q_{3L}, \ q_{4L}, \ q_{5L}, \ q_{6L}, \ q_{7L}, \\
     q_{1R}, \ q_{2R}, \ q_{3R}, \ q_{4R}, \ q_{5R}, \ q_{6R}, \ q_{7R}]^T, \\
\end{split}
\label{state_vector}
\end{equation}
where, 
$(q_x \ q_y \ q_z)$ and $(q_{\textrm{yaw}} \ q_{\textrm{pitch}} \ q_{\textrm{roll}})$ are the Cartesian coordinates of the pelvis and the Euler Angles in the Z-Y-X order, and $(q_{1L},...,q_{7L})$, $(q_{1R},...,q_{7R})$ are the generalized coordinates of the left and right legs, respectively. These correspond to the DOFs for each leg and are defined in \eqref{state_matrix}.
\begin{gather}
 \begin{bmatrix} q_1 \\ q_2 \\ q_3 \\ q_4 \\ q_5 \\ q_6 \\ q_7 \end{bmatrix}
 =
  \begin{bmatrix}
   \textrm{hip roll} \\
   \textrm{hip yaw} \\
   \textrm{hip pitch} \\
   \textrm{knee pitch} \\
   \textrm{shin pitch} \\
   \textrm{tarsus pitch} \\
   \textrm{toe pitch}
   \end{bmatrix}.
   \label{state_matrix}
\end{gather}

\begin{figure}
    \centering
    \begin{subfigure}[b]{0.6\columnwidth}
        \centering
        \includegraphics[width=\textwidth]{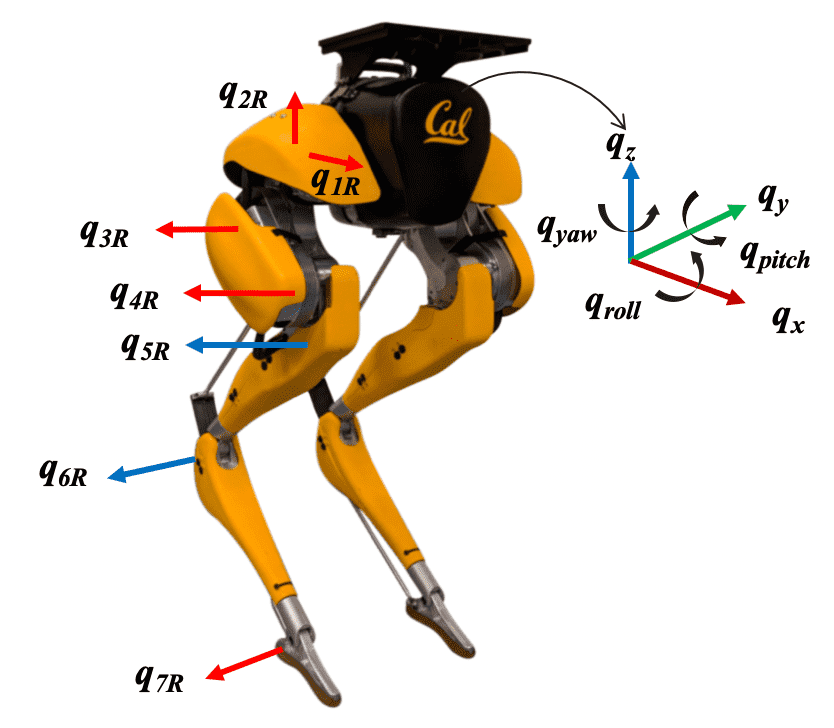}
        \caption{}
        \label{cassie_model}
    \end{subfigure}
    \begin{subfigure}[b]{0.3\columnwidth}
        \centering
        \includegraphics[width=\textwidth]{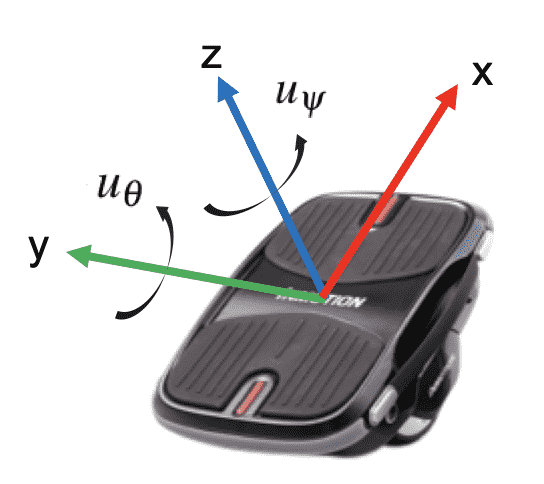}
        \caption{}
        \label{hovershoe_model}
    \end{subfigure}
    \caption{(a) Kinematic model of Cassie showing the robot's generalized coordinates in the body frame. (b) Hovershoe model where $u_\theta$ and $u_\psi$ change the pitch and yaw of the Hovershoe in the X-Y-Z body frame, respectively. }
    \vspace{-5mm}
\end{figure}

Fig. \ref{cassie_model} shows the generalized coordinates of Cassie's pelvis and right leg. The generalized coordinates of Cassie's left leg are similar to the right leg states. Each of Cassie's legs has seven DOFs with five of them being actuated: $q_1$, $q_2$, $q_3$, $q_4$, and $q_7$. The corresponding motor torques are $u_1$, $u_2$, $u_3$, $u_4$, and $u_5$. The other two DOFs, $q_5$ and $q_6$, are passive, corresponding to stiff springs. 

The dynamics of Cassie can then be expressed in the following Euler-Lagrange dynamics:
\begin{equation}
    D(q)\ddot{q} + H(q,\dot{q}) = Bu + J_s^T(q)\tau_s + J_{c}^T(q)F_{c},
    \label{lagrange}
\end{equation}
where $q$ is the generalized coordinate vector as defined in \eqref{state_vector}, $D(q)$ is the mass matrix, $H(q,\dot{q})$ contains the centripetal, Coriolis, and gravitation terms, $B$ is the motor torque matrix, $u$ is the motor torque vector of dimension $10$ corresponding to the actuators on the two legs, $J_s(q)$ is the Jacobian for the spring torques, $\tau_s$ is the spring torque vector, $J_{c}(q)$ is the Jacobian for the ground contact forces, and $F_{c}$ is the ground contact force vector.


\subsection{Dynamical Model of Hovershoes}
The Hovershoes is a highly sensitive wheeled platform. It has an internal controller that regulates the pitch to zero when there is no external force. The internal parameters and states of the hovershoe are unknown. Here, we develop a closed-loop dynamical model of the Hovershoe.
We consider $\theta$, $\psi$, $u_{\theta}$, $u_{\psi}$ to be the pitch and yaw angles and torques in the body X-Y-Z frame, and $x$, $y$, $v$ to be the x, y position and scalar speed in the global X-Y frame. Our dynamical model then is,
\begin{gather}
\label{Hovershoe1}
J_\theta \ddot{\theta} = - c_1 \theta - c_2 \dot{\theta} + u_\theta, \\
\label{Hovershoe2}
J_\psi \ddot{\psi} = - c_3 \dot \psi + u_\psi, \\
\label{Hovershoe3}
\dot{x}  = v \cos(\psi), \\
\label{Hovershoe4}
\dot{y}  = v \sin(\psi), \\
\label{Hovershoe5}
m \dot{v}  = c_4 \theta. 
\end{gather}
 
Here, the parameters ($c_1, c_2, c_3$) correspond to the Hovershoe internal controller, $c_4$ corresponds to the contact model between the Hovershoe and the ground. 
The same dynamics is used for the left and right Hovershoes.


The Hovershoe pitch dynamics is given by \eqref{Hovershoe1}, where $J_\theta$ is the moment of inertia of the Hovershoe about the y-axis, $c_1$ is the coefficient for the stiffness term, $c_2$ is the coefficient for the damping term, and $u_\theta$ is the input torque from a rider's toe about the y-axis. The ``feedback'' with $\theta$ and $\dot{\theta}$ terms in this equation account for the Hovershoe's internal stabilization controller that drives $\theta$ to zero when $u_{\theta}$ is zero. 


The yaw dynamics of the Hovershoe is described by \eqref{Hovershoe2}, where $J_\psi$ is the moment of inertia of the Hovershoe about the z-axis, $c_3$ is the coefficient for the ground contact damping term, and $u_\psi$ is the input torque from the rider's toe about the z-axis. Here, we only have a $\dot{\psi}$ term since there is no internal Hovershoe controller driving $\psi$ to zero when $u_{\psi}$ is zero. 
Figure \ref{hovershoe_model} shows $u_\theta$ and $u_\psi$ on the Hovershoe.


The global translational dynamics of the Hovershoe are captured by \eqref{Hovershoe3}-\eqref{Hovershoe4},
where $v$ is the speed and $m$ the mass of the Hovershoe.
Finally, the acceleration dynamics of the Hovershoe is captured by \eqref{Hovershoe5}
as a function of Hovershoe pitch angle.

Note that we did not conduct system identification to determine the parameters of the model. Since computing the tilt angle and applied torque on an accelerating Hovershoe was hard, we instead focused on capturing the structure of the Hovershoe model for use in our control design.

\subsection{Contact Model between Cassie and the Hovershoes}
The Cassie robot interacts with the Hovershoes through feet contacts. The contact force is $F_{c} = \begin{bmatrix}F_1, \cdots, F_4\end{bmatrix}^T$ in \eqref{lagrange} corresponding to the four contacts (front and back contact locations for either foot) with each $F_i$ having an x, y, and z components in the local Hovershoe frame, applied on Cassie. 

The two torques $u_\theta$ and $u_\psi$, with respect to y-axis and z-axis, respectively are computed as,
\begin{gather}
\begin{bmatrix}u_\theta \\ u_\psi\end{bmatrix} = \begin{bmatrix}0 & 1 & 0 \\ 0 & 0 & 1 \end{bmatrix} ~~ \sum_{i=1}^{4} r_i \times -F_i,
\label{gcm1}
\end{gather}
where,  $r_i$ is the relative position of the $i^{th}$ contact locations with respect to each Hovershoe position, and i = \{1,2\} for left hovershoe and i = \{3,4\} for right hovershoe. 
The contact forces $F_i$ are themselves determined through a compliant ground model with stick-slip friction and are then used in $F_c$ in \eqref{lagrange}, while $u_\theta, u_\psi$ from \eqref{gcm1} are used in \eqref{Hovershoe1}-\eqref{Hovershoe2}.

\section{Control Design}
\label{sec:Control}

\begin{figure}
    \centering
    \includegraphics[width=0.9\columnwidth]{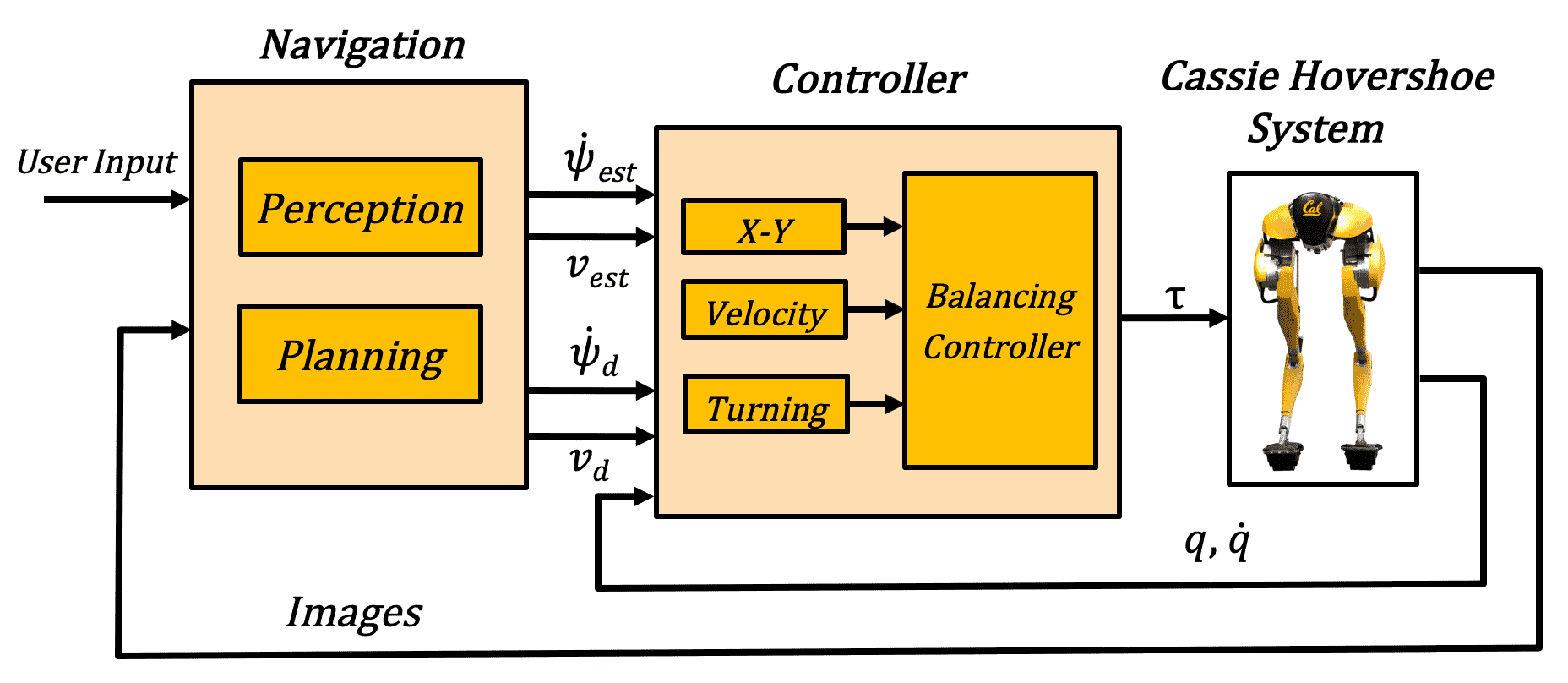}
    \caption{Autonomy framework where $\dot{\psi}_{est}$ is yaw rate of change estimated from VIO, $v_{est}$ is the estimated velocity from VIO, $\dot{\psi}_{d}$ is the desired yaw rate of change given from the path planner, and $v_{d}$ is the desired velocity given from the path planner. The user provides a goal location as input.
    }
    \label{control_diagram}
    \vspace{-5mm}
\end{figure}

Having presented the dynamical model of the Cassie with Hovershoes system in Section \ref{sec:Dynamics}, we now proceed to discuss our control strategy that will be implemented on Cassie in order to achieve our goal of driving the Hovershoes to desired locations or at desired speeds while Cassie still balances on them. It must be noted that we can neither directly measure the Hovershoes internal states nor can we directly command input torques on the Hovershoe. Our only means of interaction between the Cassie and Hovershoe are through the contact forces.
Fig. \ref{control_diagram} illustrates our autonomy framework detailing the interactions between the vision, planner, controller, and the Cassie with Hovershoes system. 

Our proposed control strategy includes a Hovershoe X-Y controller, a velocity controller, and a turning controller, all integrated with a balancing controller. The X-Y controller regulates the relative position of each Hovershoe with respect to Cassie. The velocity controller tracks a desired translational velocity and the turning controller maintains a desired rotational velocity. The balancing controller incorporates the output of the three controllers and regulates the position of Cassie's COM. 
%
Note that the gains of our proposed controller are empirically determined.

\subsection{Hovershoe X-Y Controller}
Each Hovershoe serves as a highly dynamic, moving platform for each foot of Cassie. A low-level controller that regulates the relative x-y position of each Hovershoe was designed for Cassie to keep the Hovershoes together. In the absence of this controller, the Hovershoes drift apart resulting in Cassie falling off of the moving Hovershoes.

The Hovershoe X-Y controller is also critical since it makes the system robust to variances that occur when the Hovershoes are initialized with a non-zero relative translation or orientation, any off-center placement of Cassie's feet on the Hovershoes, and any dynamic variances between the Hovershoes.

The x-axis controller keeps the relative x-coordinate of the Hovershoes at zero in the local Cassie frame.  If one of the Hovershoes is ahead of the other, then Cassie needs to bring one Hovershoe back and the other forward by pitching the Hovershoe accordingly with its toe (recall from \eqref{Hovershoe5} that increasing the pitch of the Hovershoe results in forward motion).
This is accomplished through an inner-outer loop controller. 
The outer loop computes the desired difference in the pitch of either toe, see \eqref{x1}-\eqref{x2}, while the inner loop computes the desired toe pitch torque to realize the toe pitch difference, see \eqref{x3}-\eqref{x5}.  Here $x_{\textrm{foot\ }L}$ and $q_{7L}$ are the left foot x-position and toe pitch angles respectively.  A similar notation is used for the right leg.
%
The computed toe pitch torque gets added to the left 
and subtracted from the right 
toe torques that will be 
specified by the nominal balancing controller.
\begin{gather}
    \label{x1}
    q_{\Delta \textrm{toe}}^{\textrm{des}} = -K_p^x (x_{\textrm{foot\ }L} - x_{\textrm{foot\ }R}),\\
    \label{x2}
    q_{\Delta \textrm{toe}} = q_{7L} - q_{7R},\\
    \label{x3}
    e_{\Delta \textrm{toe}} = q_{\Delta \textrm{toe}} - q_{\Delta \textrm{toe}}^{\textrm{des}},\\
    \label{x5}
    u_5^{\Delta \textrm{toe}} = -K_p^{\Delta \textrm{toe}}  e_{\Delta \textrm{toe}} -K_d^{\Delta \textrm{toe}}  \dot e_{\Delta \textrm{toe}}.
\end{gather}


The y-axis controller is designed to keep the y-coordinate of the left and right toe at $y_{\textrm{offset}}$ and $-y_{\textrm{offset}}$, 
respectively. Due to the Hovershoes' nonholonomic constraint, the Hovershoe can not move sideways directly along the y-axis. The y-coordinate can only be changed when the Hovershoes yaw and move forward, so that the legs can come closer or move apart depending on the yaw angle of each Hovershoe. The y-axis controller has an inner-outer loop as well. The outer loop determines each hip's desired yaw angle through the difference between each toe's y-coordinate and desired $y_{\textrm{offset}}$, see \eqref{y_cont1}-\eqref{y_cont2}. The hip yaw is used since the toe does not have an individual motor for toe yaw. In the inner loop, the desired hip yaw angle is converted into hip yaw motor torque, see \eqref{y_cont4}. 
\begin{gather}
    \label{y_cont1}
    e_{y} = y_{\textrm{foot}} \pm y_{\textrm{offset}}, \\ 
    \label{y_cont2}
    q_{2}^{\textrm{des}} = -K_p^y ~ e_{y}, \\
    \label{y_cont4}
    u_2^{y} = - K_p^{q_2} ~ (q_{2} - q_{2}^{\textrm{des}}).
\end{gather}
In the above equations, $\pm$ refers to addition to one leg and subtraction for the other.
To achieve a wave pattern with the Hovershoes, we can specify a time-varying $y_{\textrm{offset}}(t)$. 


\subsection{Velocity Controller}

Once Cassie can balance on the Hovershoes and maintain a relative position between the Hovershoes while they are moving, we can then design higher-level controllers for Cassie to perform speed regulation and turning maneuvers. These high-level controllers are required for interfacing with the path planner in the autonomous system.

The velocity controller maintains the velocity of Cassie on the Hovershoes at a specified set point given either by an operator or by the path planner. Velocity control is accomplished through regulating the $x$-COM position of Cassie
, where the desired $x$-COM position is proportional to the error in velocity:
\begin{equation}
    \label{vel3}
    COM_{x}^{\textrm{des}} = - K_p^{\textrm{vel}} ~ (v_{est} - v_d) 
\end{equation}
 Here, 
 $v_{est}$ is the velocity of Cassie estimated from our vision system and $v_d$ is the desired velocity.
 
\subsection{Turning Controller}
The turning controller maintains a desired yaw rate for Cassie's pelvis that is specified either by an operator or by the path planner. For the two Hovershoes to turn along a curve, the Hovershoes not only need to constantly yaw but the outer Hovershoe needs to go faster than the inner Hovershoe.  Furthermore, for fast turns, Cassie will also need to bank by leaning into the turn.  Leaning into a turn ensures Cassie remains balanced on the Hovershoes during fast turning maneuvers and does not tip over.
This is accomplished by a three-part controller.
%
The Hovershoes are turned by yawing Cassie's hips proportional to the yaw rate error, see \eqref{turn1}-\eqref{turn2}.
The outer Hovershoe is made to go faster than the inner Hovershoe by differentially actuating Cassie's toes, see \eqref{turn3} where the computed toe difference torque will be added to one toe and subtracted from the other in the balancing controller.
%
Finally, Cassie is made to lean into the turn by changing the desired COM position along the y-axis when there is a large angular and/or tangential velocity,  see \eqref{turn4}-\eqref{turn5}. 
\begin{gather}
    \label{turn1}
    e_{\dot{\psi}}
    = \dot{\psi}_{est} - \dot{\psi}_d, 
    \\
    \label{turn2}
    u_{2}^{\textrm{turn}} = -K_p^{\textrm{yaw}} e_{\dot{\psi}}, \\ 
    \label{turn3}
    u_{5}^{\textrm{turn}} = -K_p^{\textrm{pitch}} e_{\dot{\psi}}, \\ 
    \label{turn4}
    \theta_{\textrm{tilt}} = \arctan{\frac{\dot{\psi}_{est} v_{est}}{9.81}}
    \\
    \label{turn5}
    COM_{y}^{\textrm{des}} = -K_p^{\textrm{shift}}(L~\theta_{\textrm{tilt}}).
\end{gather}
 Here, $\dot{yaw}_{\textrm{vision}}$ is the actual velocity of Cassie computed from our vision system and $L$ is the distance of Cassie's COM to the Hovershoes platform plane.

\subsection{Integrating with a Nominal Balancing Controller}

Our proposed controller is based on a nominal balancing controller for Cassie from \cite{c9}. The nominal balancing controller works by 
regulating the COM position error and the torso orientation. In order to move the COM along the x-axis, the controller modules the toe pitch torque. In order to move the COM along the y-axis, the controller changes the leg length difference between the left and right legs.  We modify the nominal torque $\tilde u_i$ from the nominal balancing controller as follows:
\begin{figure}
    \centering
    \includegraphics[width=0.6\columnwidth]{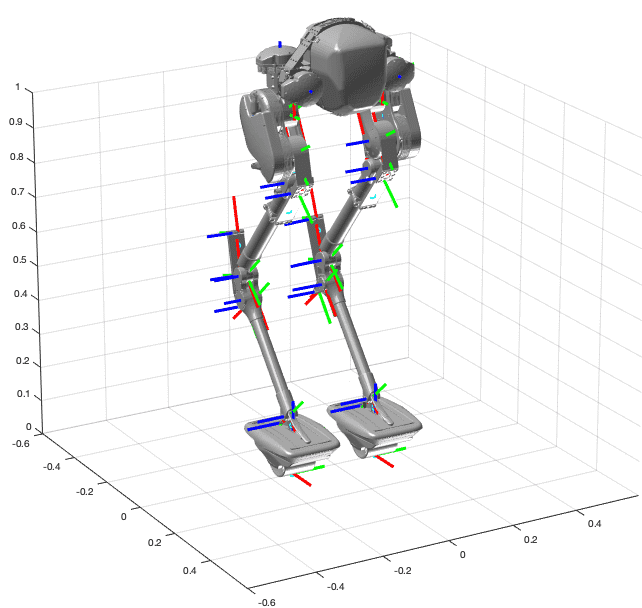}
    \caption{Simulation visualization with Cassie on the Hovershoes. This is a static balancing task demonstration.}
    \label{featherstone_cassie}
    \vspace{-1mm}
\end{figure}

\begin{figure}
    \centering
    \begin{subfigure}[t]{0.49\columnwidth}
        \centering
        \includegraphics[trim=0 200 0 200, clip, width=1\columnwidth]{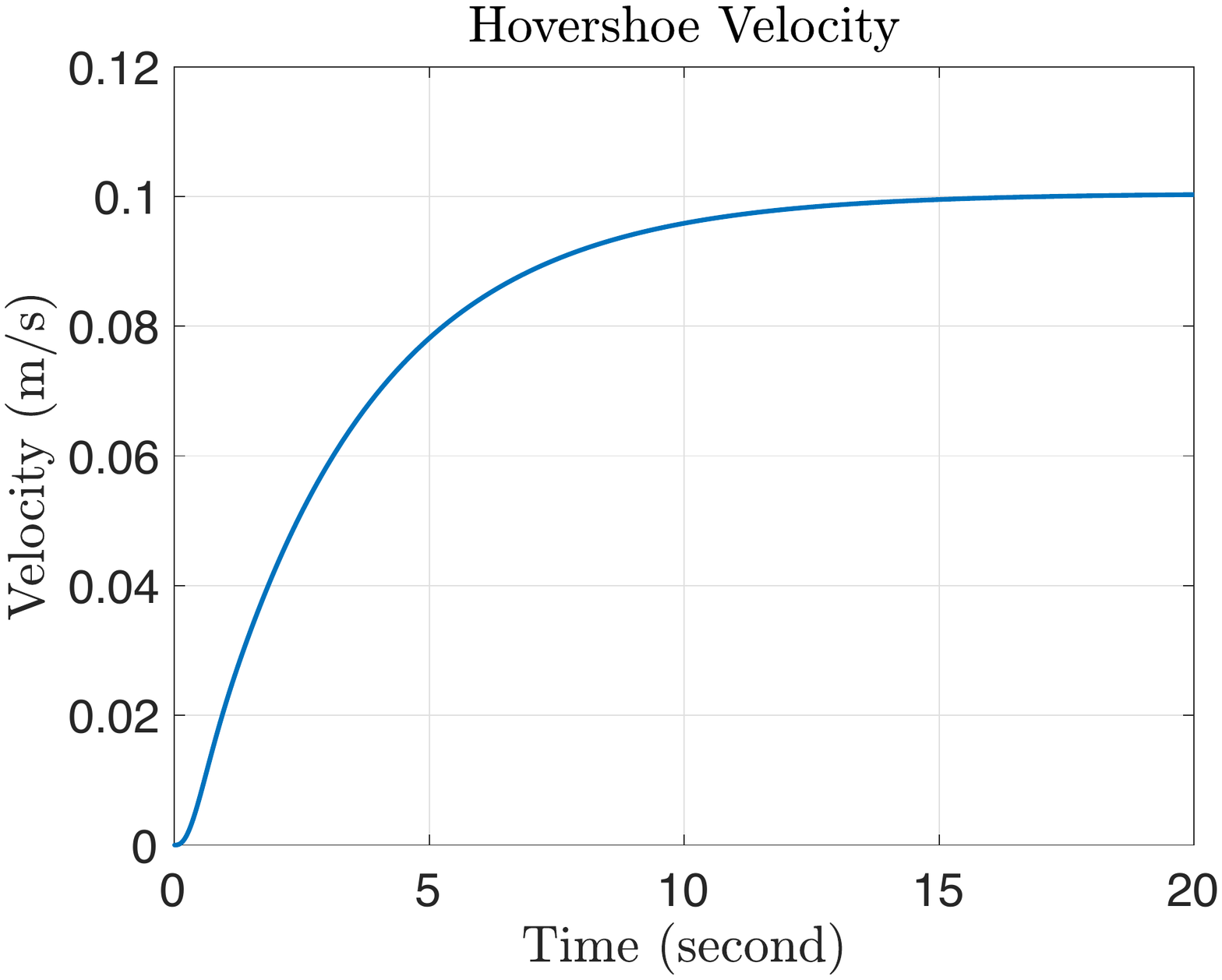}
        \caption{}
        \label{sim_vel}
    \end{subfigure}
    \begin{subfigure}[t]{0.49\columnwidth}
        \centering
        \includegraphics[trim=0 200 0 200, clip, width=1\columnwidth]{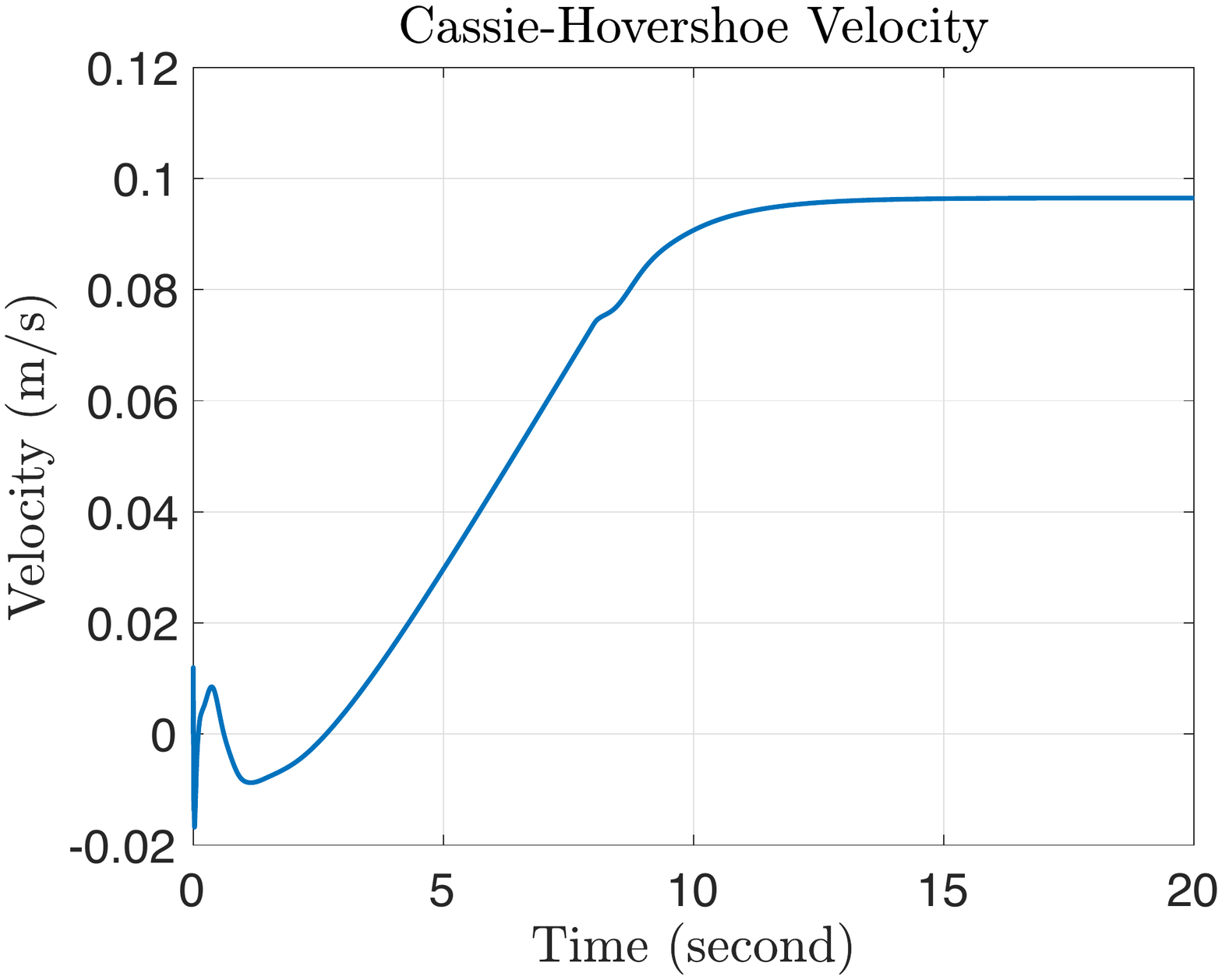}
        \caption{}
        \label{featherstone_vel_real}
    \end{subfigure}
    \begin{subfigure}[t]{0.49\columnwidth}
        \centering
        \includegraphics[trim=0 200 0 200, clip, width=1\columnwidth]{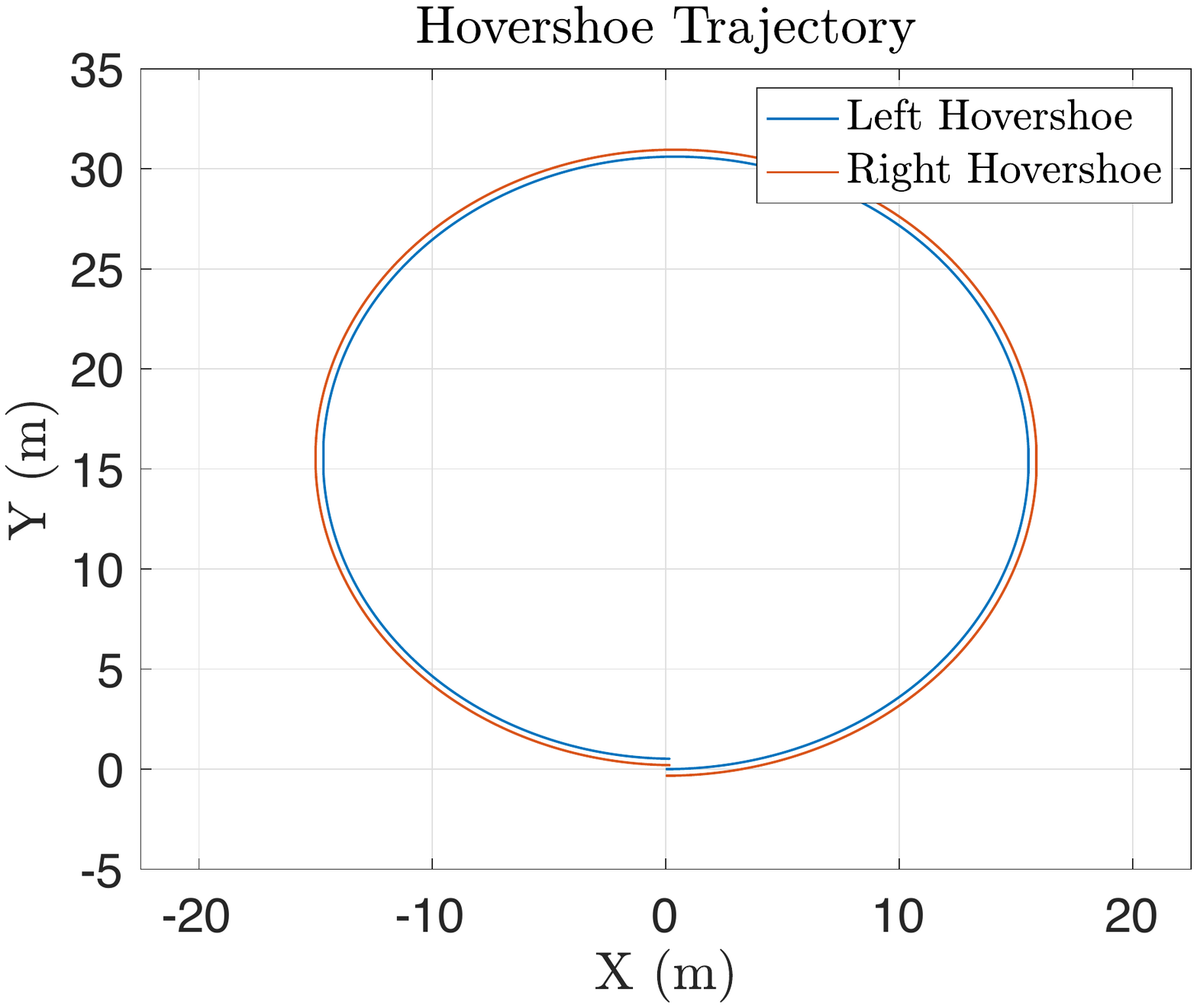}
        \caption{}
        \label{sim_turn_real}
    \end{subfigure}   
    \begin{subfigure}[t]{0.49\columnwidth}
        \centering
        \includegraphics[trim=0 200 0 200, clip, width=1\columnwidth]{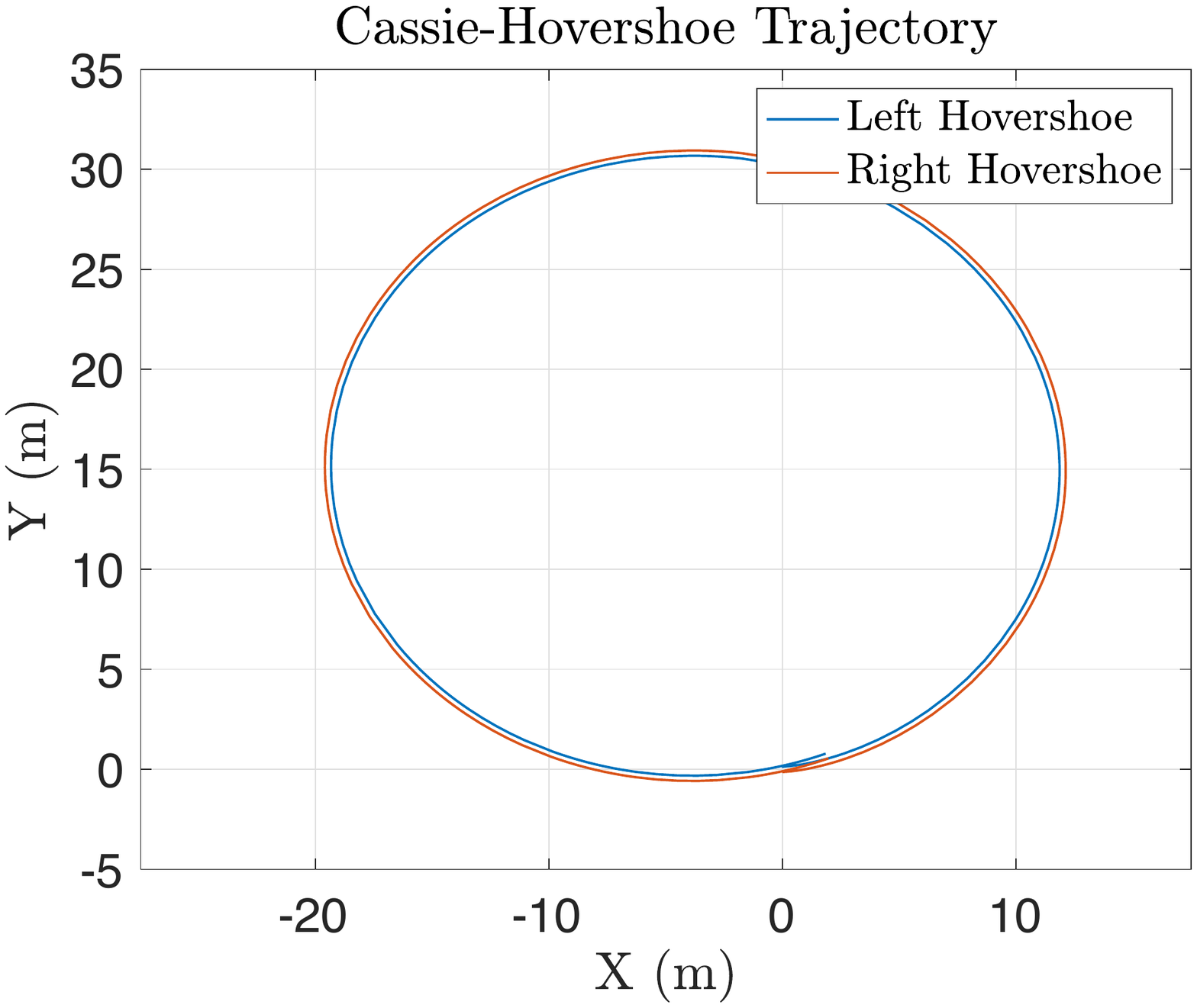}
        \caption{}
        \label{featherstone_turn_real}
    \end{subfigure} 
    \begin{subfigure}[t]{0.49\columnwidth}
        \centering
        \includegraphics[trim=0 200 0 200, clip, width=1\columnwidth]{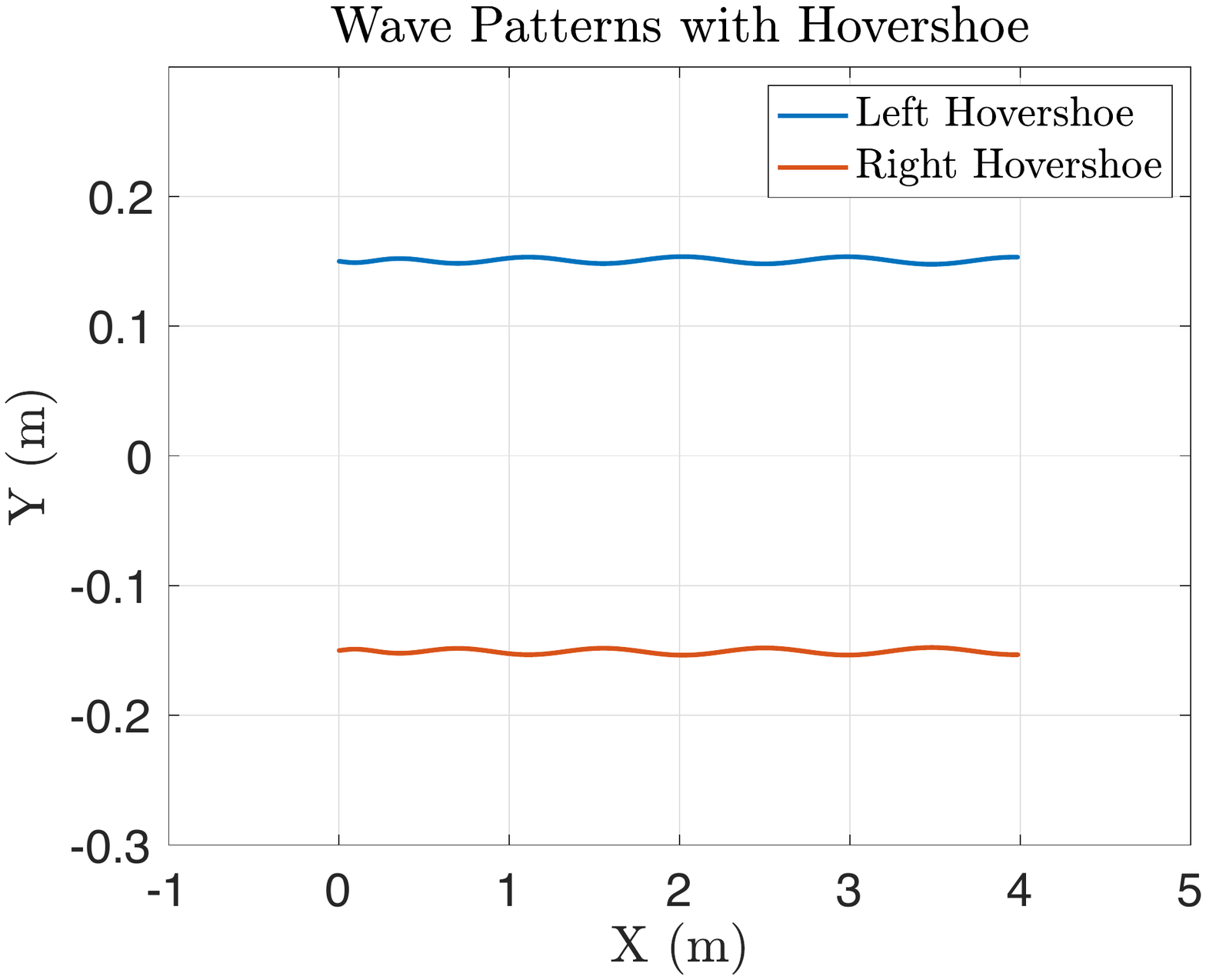}
        \caption{}
        \label{sim_wave}
    \end{subfigure} 
    \begin{subfigure}[t]{0.49\columnwidth}
        \centering
        \includegraphics[trim=0 200 0 200, clip, width=1\columnwidth]{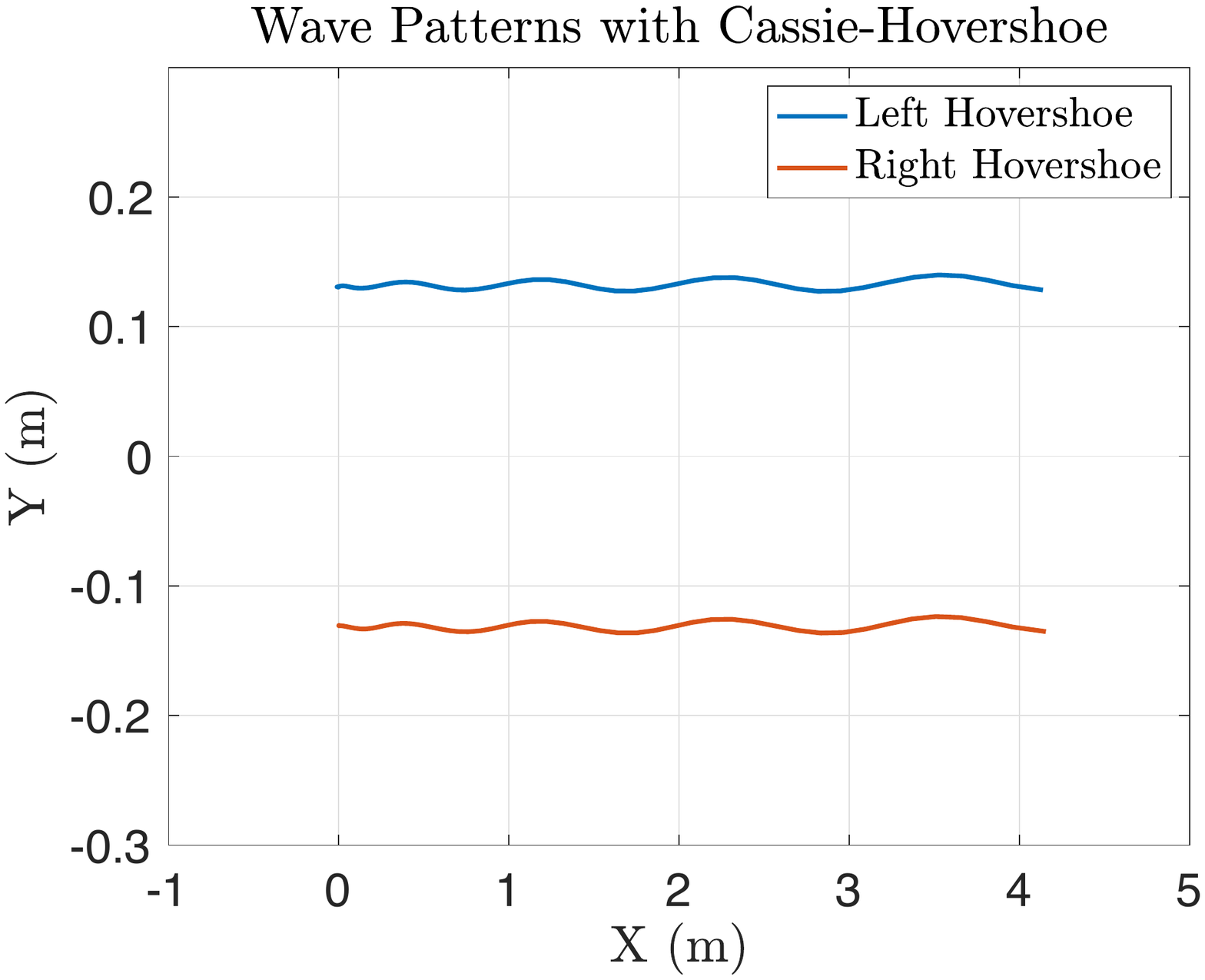}
        \caption{}
        \label{featherstone_wave_real}
    \end{subfigure} 
    \caption{Simulation results for the Hovershoes (left column) and Cassie-Hovershoe system (right column) for (a,b) forward velocity tracking, (c,d) going around in a circle, and (e,f) following a wave pattern.}
    \label{Hovershoe_Simulation}
    \vspace{-8mm}
\end{figure}

\begin{gather}
    \label{nominal_u1}
    u_1 = \tilde u_1, \\
    \label{nominal_u2}
    u_2 = \tilde u_2 + u_2^{y} + u_{2}^{\textrm{turn}}, \\
    \label{nominal_u3}
    u_3 = \tilde u_3(COM_{y}^{\textrm{des}}), \\
    \label{nominal_u4}
    u_4 = \tilde u_4(COM_{y}^{\textrm{des}}), \\
    \label{nominal_u5}
    u_5 = \tilde u_5(COM_{x}^{\textrm{des}}) \pm u_5^{\Delta \textrm{toe}} \pm u_5^{\textrm{turn}} + u_5^{\textrm{damping}}.
\end{gather}
In particular, the hip yaw torque $u_2$ in \eqref{nominal_u2} comprises of the yaw torque from the nominal balancing controller $\tilde u_2$, as well as torques from the y-axis controller $u_2^{y}$, and turning controller $u_{2}^{\textrm{turn}}$. 
The toe torque $u_5$ in \eqref{nominal_u5} comprises of the nominal balancing controller toe torque $\tilde u_5$ as well as the toe pitch difference torque $u_5^{\Delta \textrm{toe}}$ from the x-axis controller, the turning controller torque $u_5^{\textrm{turn}}$, and a damping controller $u_5^{\textrm{damping}}$ that suppresses the oscillation of the toe pitch joint. 
As mentioned earlier, $\pm$ refers to a torque that is added to one leg and subtracted from the other.

\section{Simulation Results}
\label{sec:Simulation}

In order to verify the proposed controller design presented in Section \ref{sec:Control} on the dynamical model from Section \ref{sec:Dynamics}, we ran two numerical simulations.  First, we considered the Hovershoes (without Cassie) and provided the controllers full state knowledge of the Hovershoes and allowed the controllers to directly actuate the Hovershoes.  In the next simulation, we considered the Cassie-Hovershoes integrated system where the controller did not have access to the Hovershoe state and could only indirectly actuate the Hovershoes through Cassie.  We discuss these simulation results in the following subsections. The simulations ran in \texttt{MATLAB} \texttt{2018a} installed on a laptop (Intel Core i5, RAM 8GB).

\begin{figure}
    \centering
    \begin{subfigure}[b]{0.45\columnwidth}
        \centering
        \includegraphics[width=\textwidth]{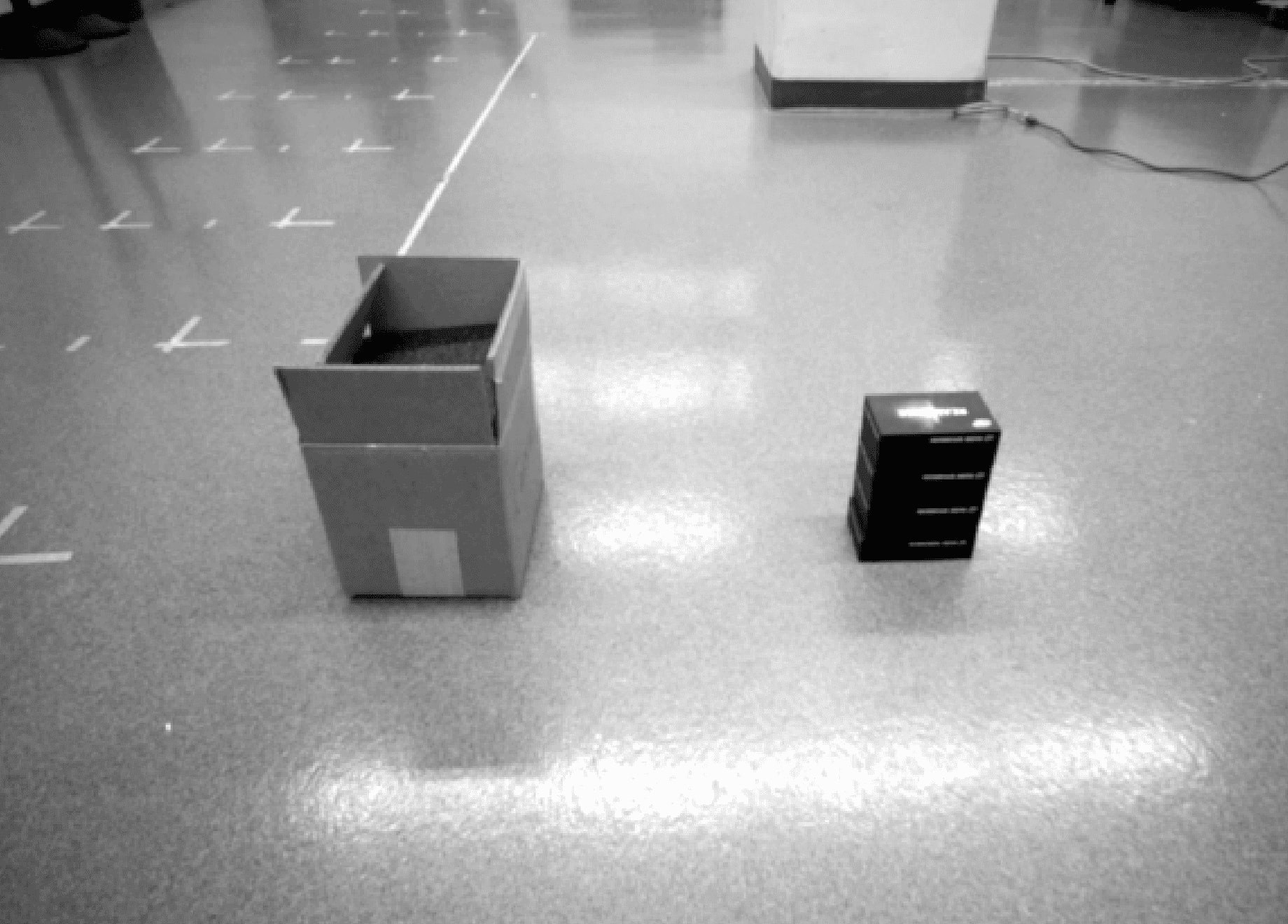}
        \label{segmentation_raw}
    \end{subfigure}
    \begin{subfigure}[b]{0.45\columnwidth}
        \centering
        \includegraphics[width=\textwidth]{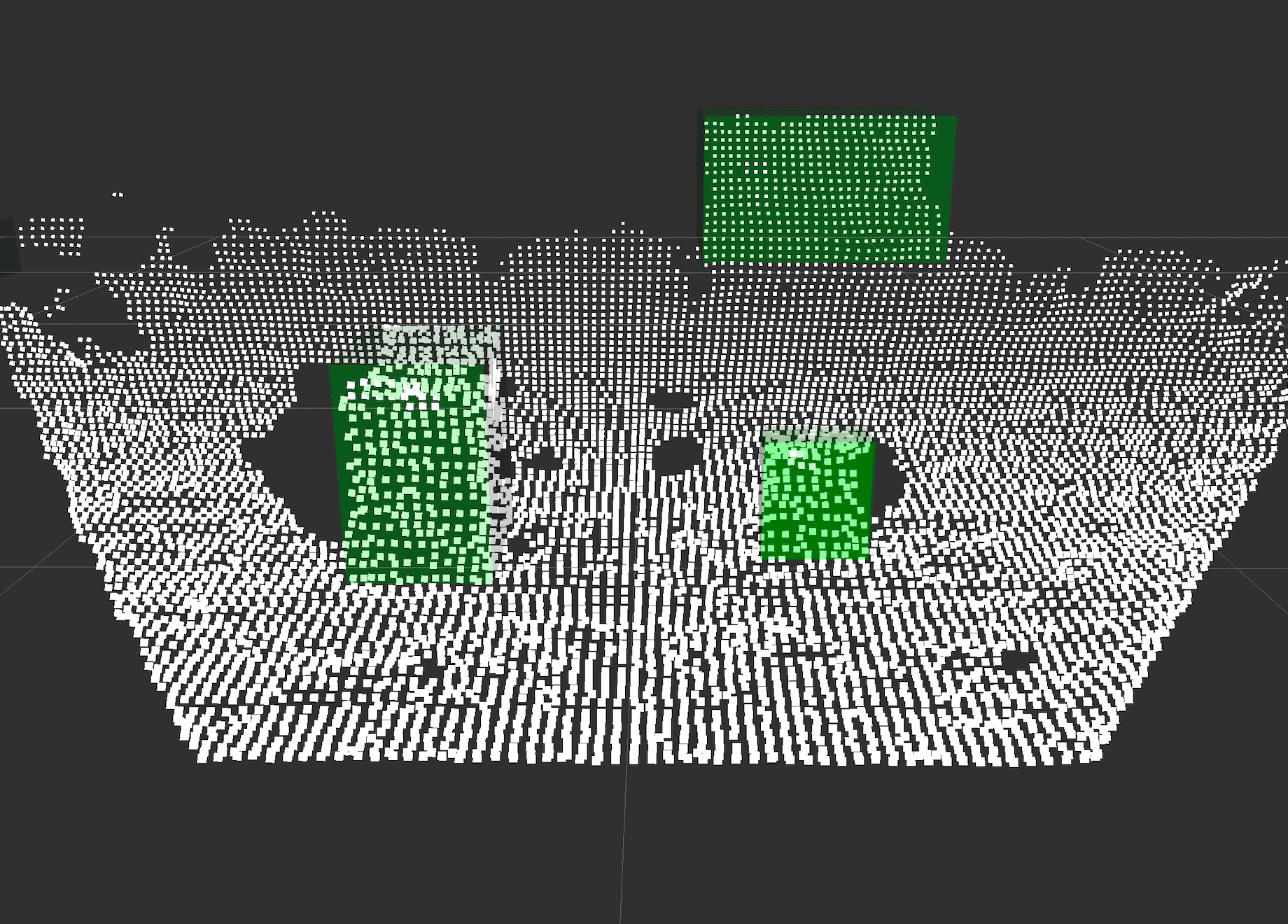}
    \label{segmentation}
    \end{subfigure}
    \caption{The left picture shows the real scene and the right demonstrates the segmented ground and obstacles.}
    \label{vision_segmentation}
    \vspace{-5mm}
\end{figure}

\subsection{Hovershoes Simulation}

We modeled each Hovershoe in Matlab with the dynamics specified in  \eqref{Hovershoe1}-\eqref{Hovershoe5}. 
In the simulation, we assumed perfect state knowledge of the Hovershoes, and the controllers would provide the input torques: $u_\theta$ and $u_\psi$. 

We verified the velocity controller by setting a desired velocity of \SI{0.1}{m/s} and verifying the tracking as shown in Fig.~\ref{sim_vel}.
We verified the turning controller by setting a desired rotational velocity and ensuring the Hovershoes turned in a circle as shown in Fig.~\ref{sim_turn_real}. Finally, we verified the wave pattern controller by setting the $y_{\textrm{offset}}$ to a sinusoidal wave, resulting in Fig.~\ref{sim_wave}.

\subsection{Cassie-Hovershoes System Simulation}

We next simulated the Cassie-Hovershoes integrated system dynamics \eqref{lagrange}-\eqref{Hovershoe5},  with our designed controller. 
Fig. \ref{featherstone_cassie} shows the visualization of Cassie and the Hovershoes balancing in the simulation. 
We used the same set points as the previous Hovershoes simulation to test our velocity, turning, and wave pattern controller and obtained a similar performance, see Figs. \ref{featherstone_vel_real}, \ref{featherstone_turn_real}, and \ref{featherstone_wave_real}. 

\begin{figure}
    \centering
    \includegraphics[width=1.0\columnwidth]{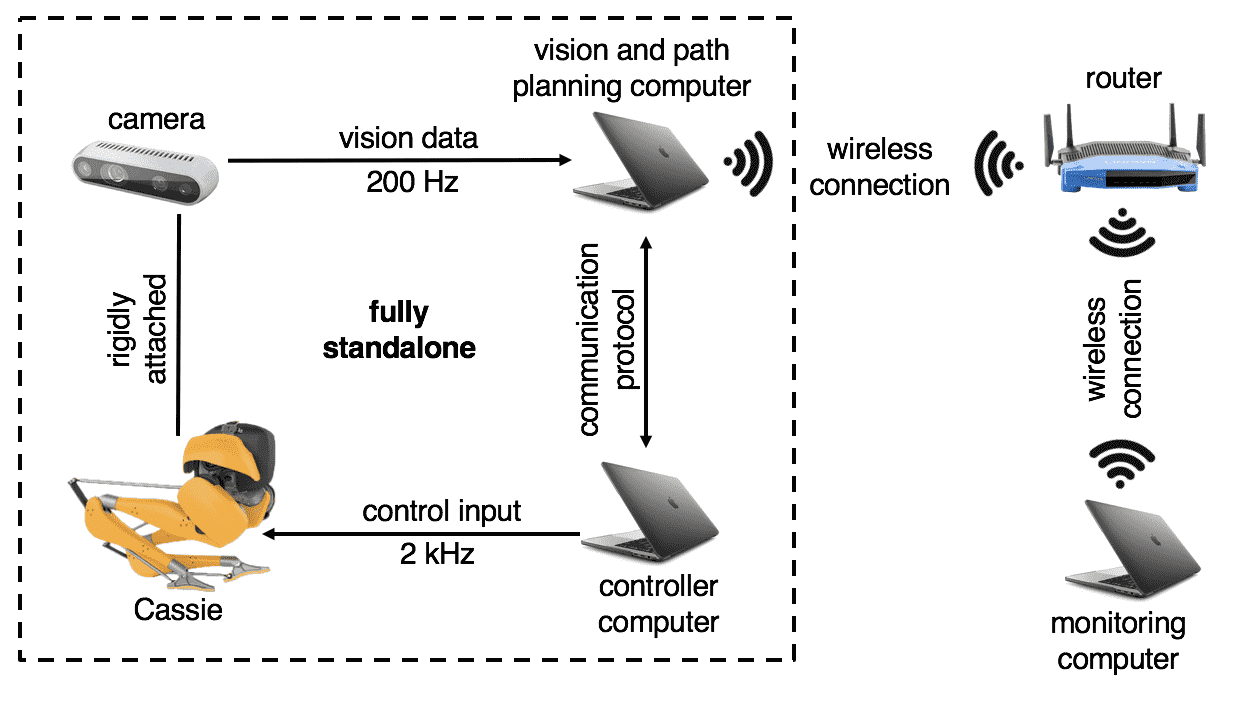} 
    \caption{Hardware infrastructure diagram for the autonomous Cassie system. The vision and path planning computer and the controller computer are onboard Cassie, while the monitoring computer is connected wirelessly and used to see signals and tune parameters.}
    \label{hardware}
    \vspace{-4mm}
\end{figure}

\section{Perception and Planning}
\label{sec:Experimental_Setup}

Having numerically validated our control strategy, we will next pave the path for experiments by presenting our perception and path planning strategies and also detailing the communication between perception, planning, and control. 
\subsection{Perception}

In order to modularize our set-up to be simple and easy to manage, and at the same time decrease the computational load, we opted for an integrated VIO sensor, the Intel RealSense T265, which is comprised of a pair of fish-eye lenses is combined with an on-board implementation and extension of an algorithm utilizing stereo VIO with loop-closure detection to produce reliable odometry \cite{c19}.
Additionally, a rigidly attached depth camera, Intel RealSense D435i, was used for the obstacle mapping. The D435i camera processed and matched the stereo images using dedicated on-board hardware. By designing the vision set-up as above, we are able to incorporate the entire perception, planning, and control pipeline using the CPUs on Cassie, and avoid the need of power-hungry GPUs.

The transformation between the two cameras was known and published internally via ROS. The depth camera was slightly tilted towards the ground to cover more ground space near Cassie. The depth maps were converted to point clouds, which were then filtered and voxelized to a horizontal resolution of 5cm and vertical resolution of \SI{10}{cm} to reduce the memory usage and accelerate the computation.
After transforming the point clouds to the base frame of the Hovershoes, a ground segmentation with a tuned tolerance to noise was applied to remove the ground plane and register the obstacles onto an global occupancy map. Fig.~\ref{vision_segmentation} shows an example of our segmentation. The map registration was achieved by using the RTAB-Map ROS package, but the built-in appearance-based loop closure detector was turned off. In order to increase the refresh rate for the local costmap, the segmented 3-D obstacles were further converted and merged to a pseudo-laser scan to decrease the data rate. Based on the above perception infrastructure, we next detail how we can achieve real-time planning in order to circumvent obstacles.


\subsection{Motion Planning}
To create real-time motion plans for the Cassie-Hovershoes system, we require the planner to be reliable and kinematically feasible. For the global planner, we employed a search-based method, Dijkstra's algorithm, for its guarantee of a solution. The costmap decay rate was tuned to confine the Dijkstra's behavior to follow a path that stays in the middle of two potentially incoming obstacles.

For the local planner, the optimization-based Timed Elastic Band (TEB) planner was applied on top of the Dijkstra's algorithm as the TEB planner looks \SI{1.75}{m} ahead of the current position and generates a short-term, kinematically feasible path for Cassie. In order to emulate the forward non-holonomic property of the Hovershoes, customized car-like optimization weights were used to encourage forward motion. The optimization process was adjusted to adapt the computing power of Cassie and the resulting planner frequency could reach up to slightly less than \SI{50}{Hz}. However, \SI{10}{Hz} proved to be sufficient for Cassie to autonomously navigate through a narrow corridor with obstacles.


\subsection{Communication between Perception, Planning \& Control}

Fig. \ref{hardware} shows a diagram of the communication setup and hardware infrastructure. Please note that the router can be replaced with a Wi-Fi hotspot. In addition, between the real-time controller and vision system, we use the UDP to communicate to each other. Finally, the vision and path planning have a separate computer other than the controller because only the controller requires a real-time computer. We use a gantry during experiments for safety purposes only.

\begin{figure}
    \centering
    \begin{subfigure}[h]{0.24\linewidth}
        \centering
        \includegraphics[width=1.0\linewidth]{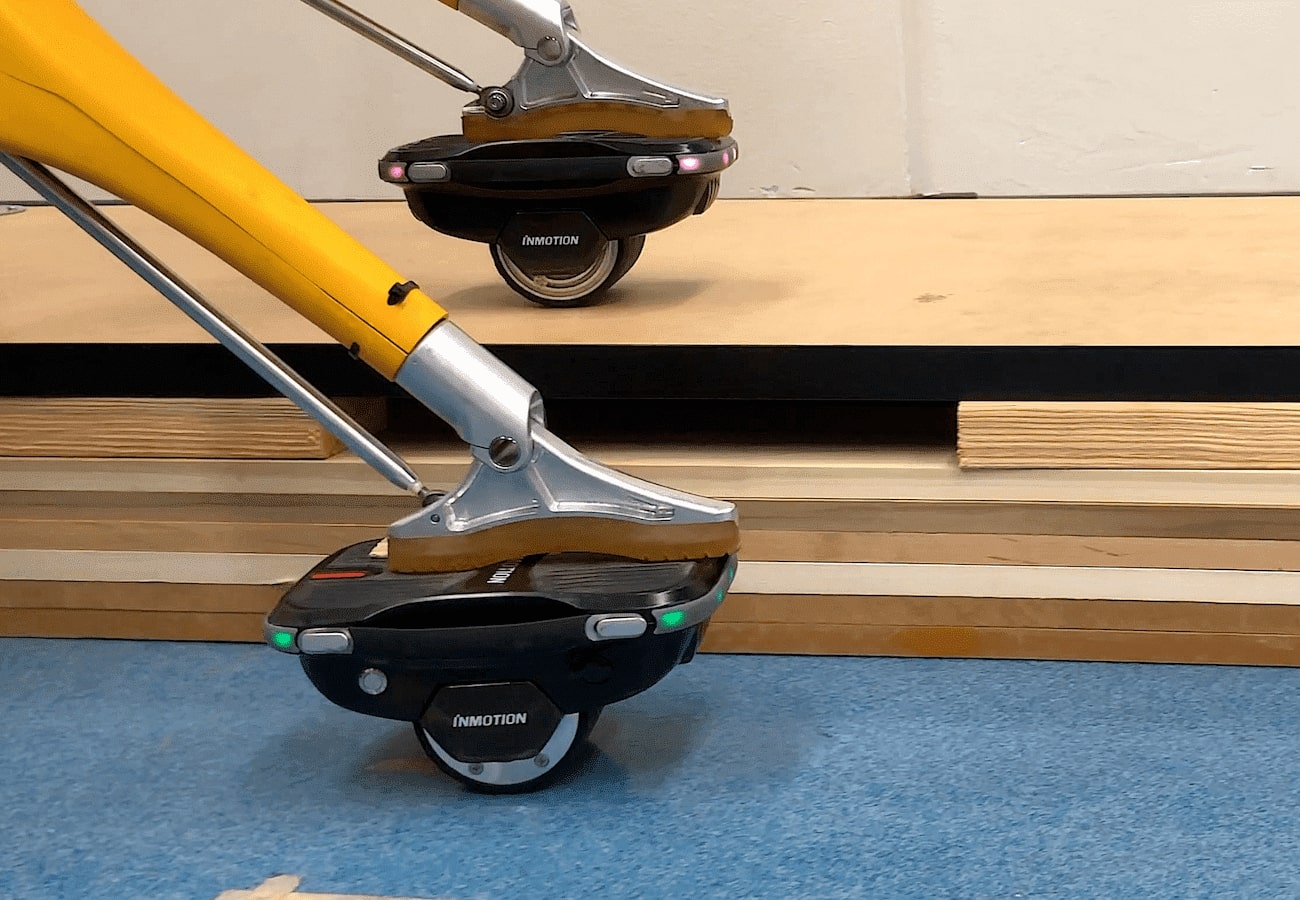}
        \caption{}
    \end{subfigure} 
    \begin{subfigure}[h]{0.24\linewidth}
        \centering
        \includegraphics[width=1.0\linewidth]{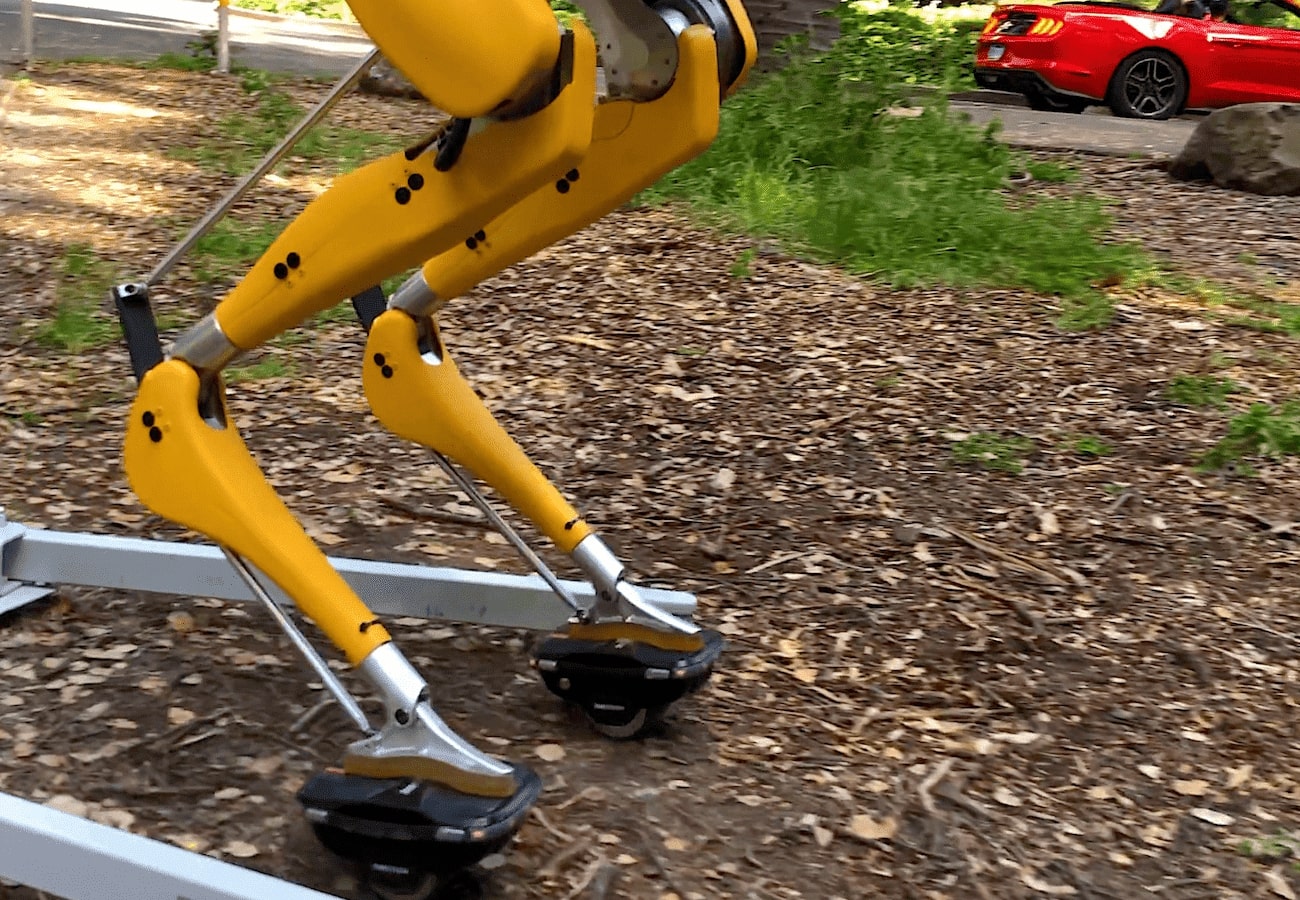}
        \caption{}
    \end{subfigure} 
    \begin{subfigure}[h]{0.24\linewidth}
        \centering
        \includegraphics[width=1.0\linewidth]{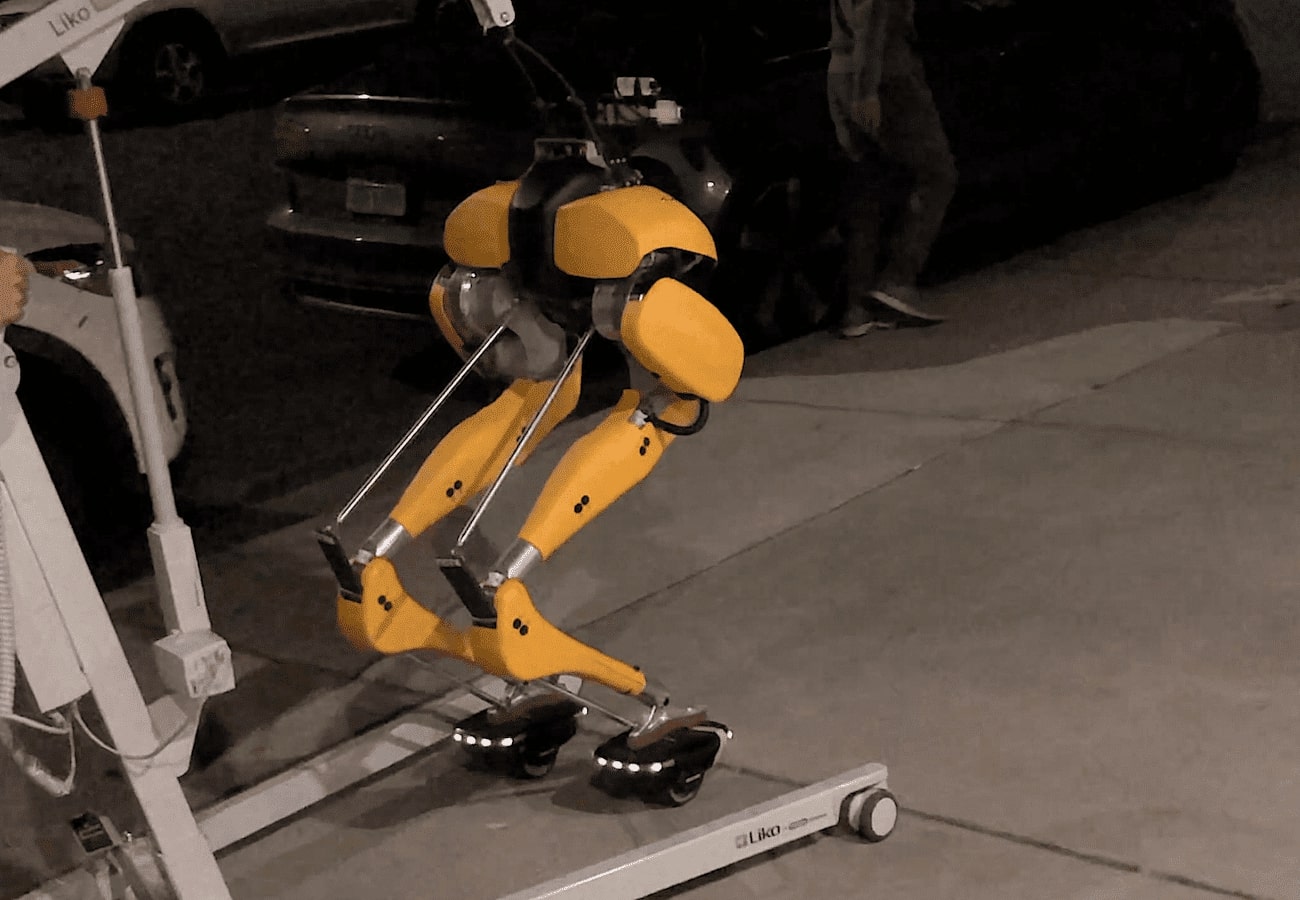}
        \caption{}
    \end{subfigure}
    \begin{subfigure}[h]{0.24\linewidth}
        \centering
        \includegraphics[width=1.0\linewidth]{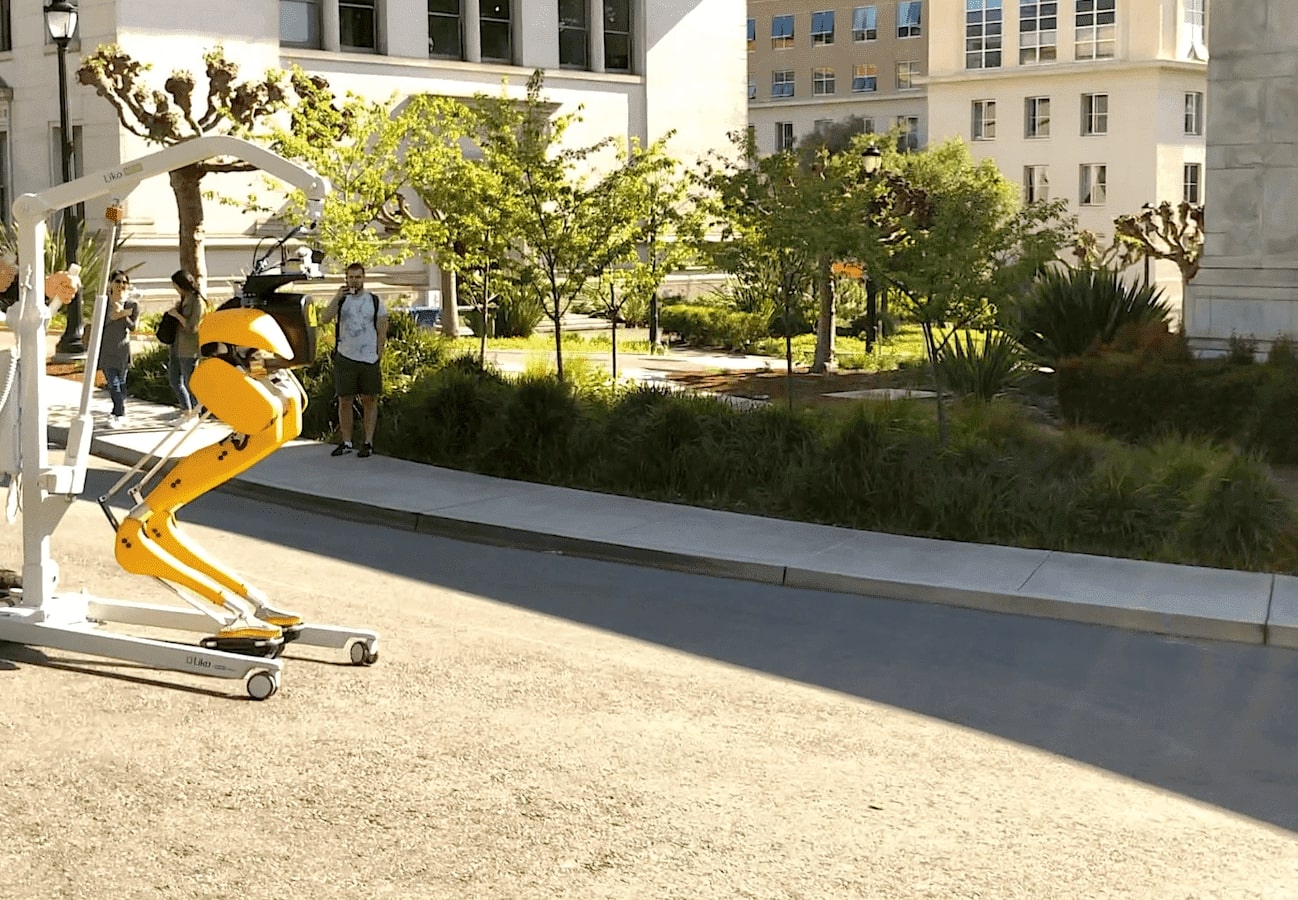}
        \caption{}
    \end{subfigure}
    \caption{Cassie riding the Hovershoes on (a) different heights, (b) rough terrain, (c) up a $7^\circ$ incline, and (d) down a $13^\circ$ incline.}
    \label{additional_skills}
\end{figure}

\begin{figure}
    \centering
    \begin{subfigure}[h]{0.32\linewidth}
        \centering
        \includegraphics[width=1.0\linewidth]{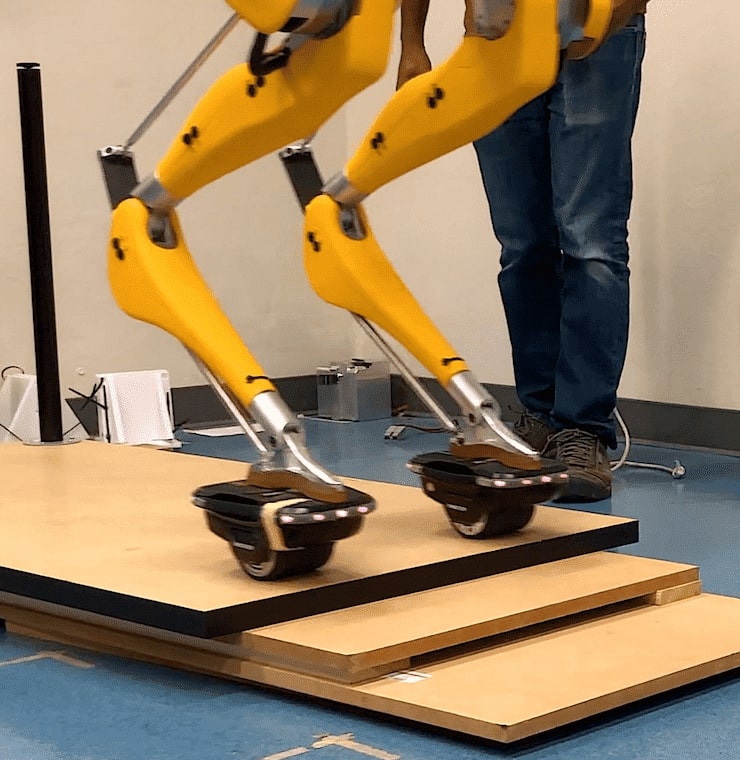}
    \end{subfigure}
    \begin{subfigure}[h]{0.32\linewidth}
        \centering
        \includegraphics[width=1.0\linewidth]{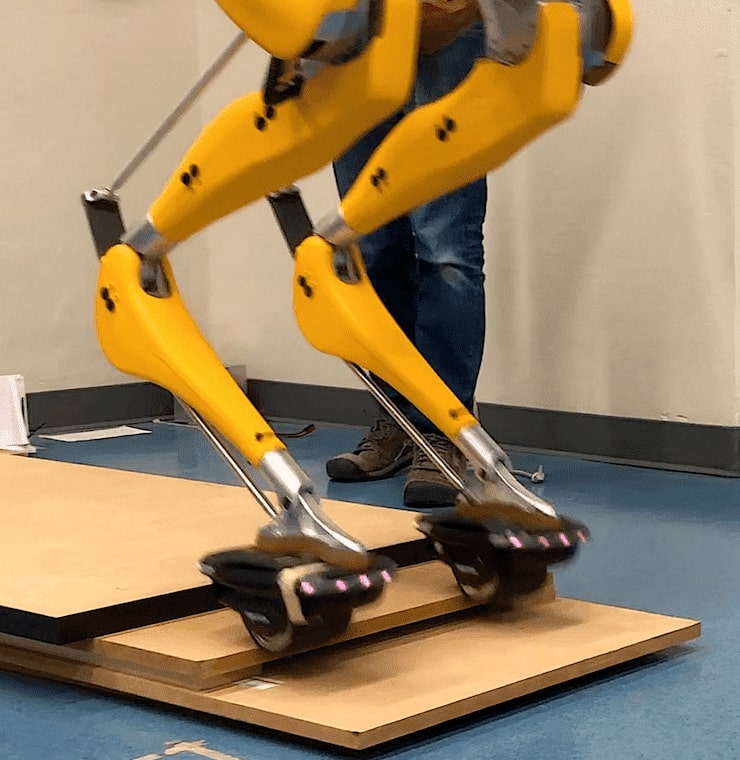}
    \end{subfigure}
    \begin{subfigure}[h]{0.32\linewidth}
        \centering
        \includegraphics[width=1.0\linewidth]{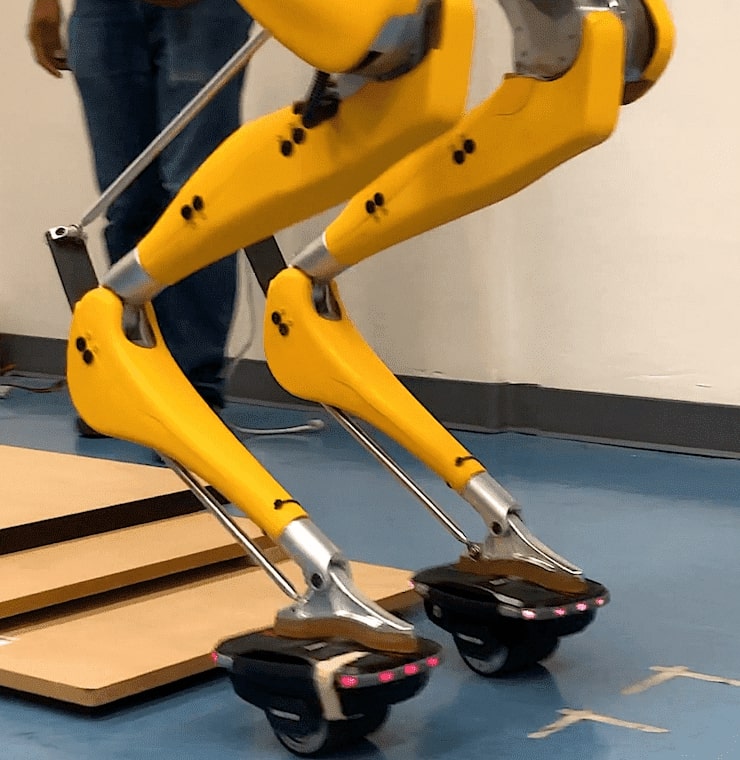}
    \end{subfigure} 
    \caption{Cassie robustly descending three stairs with heights $79\%$, $93\%$, $93\%$ of the wheel radius respectively.}
    \label{robustness}
    \vspace{-4mm}
\end{figure}

\begin{figure}
    \centering
    \includegraphics[trim=0 200 0 200, clip, width=1.0\columnwidth]{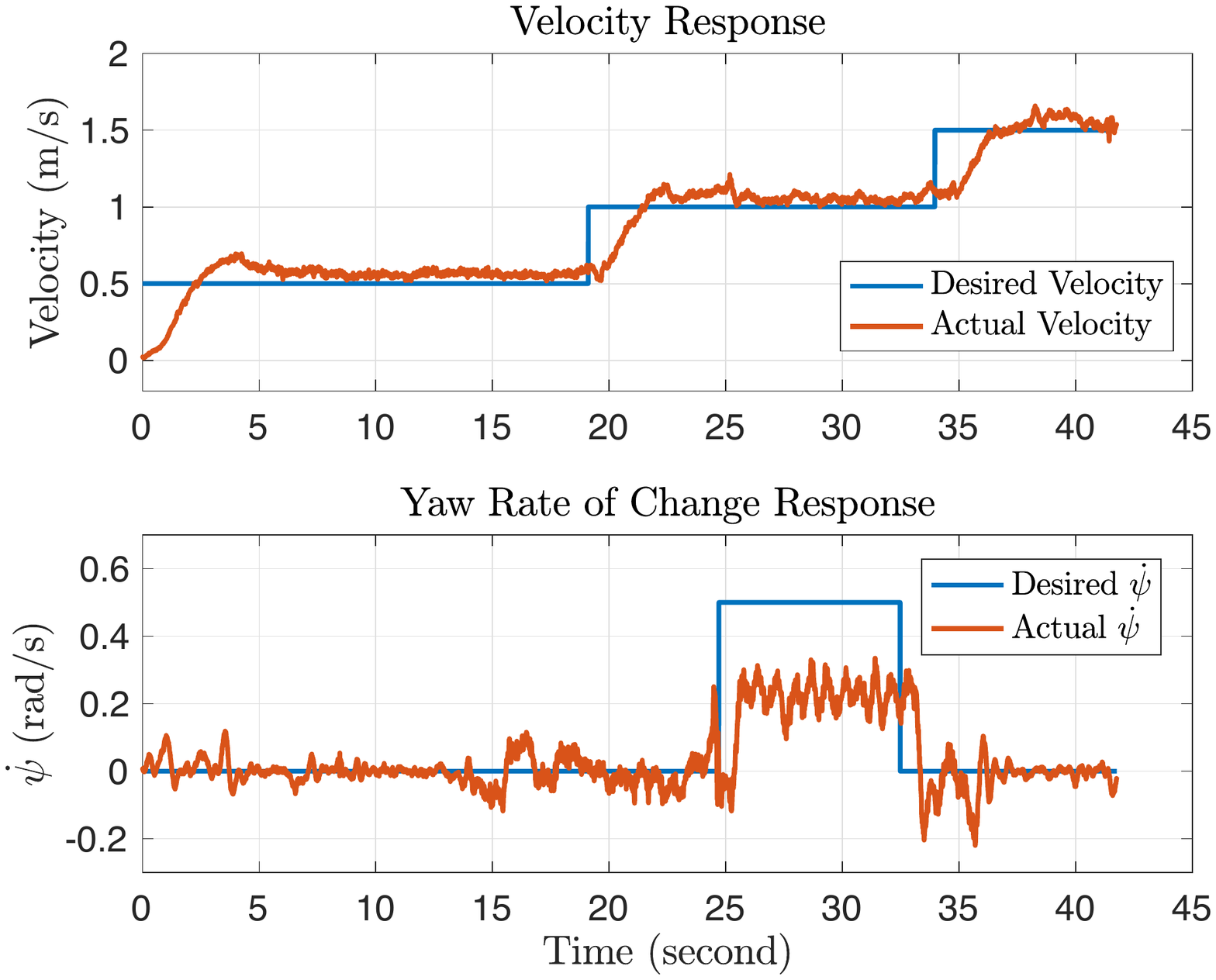}
    \caption{Top: velocity response to three successive step inputs; Bottom: yaw rate of change response to two successive step inputs.}
    \label{cassie_vel}
    \vspace{-4mm}
\end{figure}

\section{Experimental Results}
\label{sec:Experimental_Results}

After developing our control strategy, path planning algorithm, and perception system, we now present the results of our autonomous framework. First, we demonstrate the robustness of our controller. Then, we show the performance of our velocity and turning maneuver. Last, we demonstrate that Cassie successfully navigating an obstacle course autonomously while riding the Hovershoes.

\subsection{Robustness}
Once Cassie could balance and perform the tasks in simulation, we transferred the controller to Cassie and checked the robustness of the controller subject to disturbances that could cause Cassie to fall. Robustness was checked by kicking the Hovershoes and pushing Cassie's pelvis and tarsus while it was riding the Hovershoes. In addition, Cassie successfully traversed over rough terrain with cracks and bumps, as well as riding on up and down slopes with $7^\circ$ and $13^\circ$ inclines respectively, and riding with Hovershoes at different heights (see Fig.~\ref{additional_skills}). A challenge task was to go down a variety of steps. Cassie successfully went down single stairs that were $79\%$ and $171\%$ of the Hovershoe wheel radius and a sequence of three stairs that were $79\%$, $93\%$, $93\%$ of the Hovershoe wheel radius, respectively. As can be seen in Fig.~\ref{robustness}, Cassie remains balanced and the Hovershoes remain level after descending each step.

\subsection{Vision-based Velocity Tracking}
We tested the performance of our vision-based velocity controller by providing a step input change in desired velocity. We found that it can track a desired velocity with a small steady-state error. 
As can be see in Fig.~\ref{cassie_vel}, the desired velocity step inputs of \SI{0.5}{m/s}, \SI{1.0}{m/s}, and \SI{1.5}{m/s}, were tracked with small steady-state errors with the root mean square error (RMSE) being \SI{0.1467}{m/s}.

\subsection{Fast Turning}
\begin{figure}
    \centering
    \begin{subfigure}[h]{0.32\linewidth}
        \centering
        \includegraphics[width=1.0\linewidth]{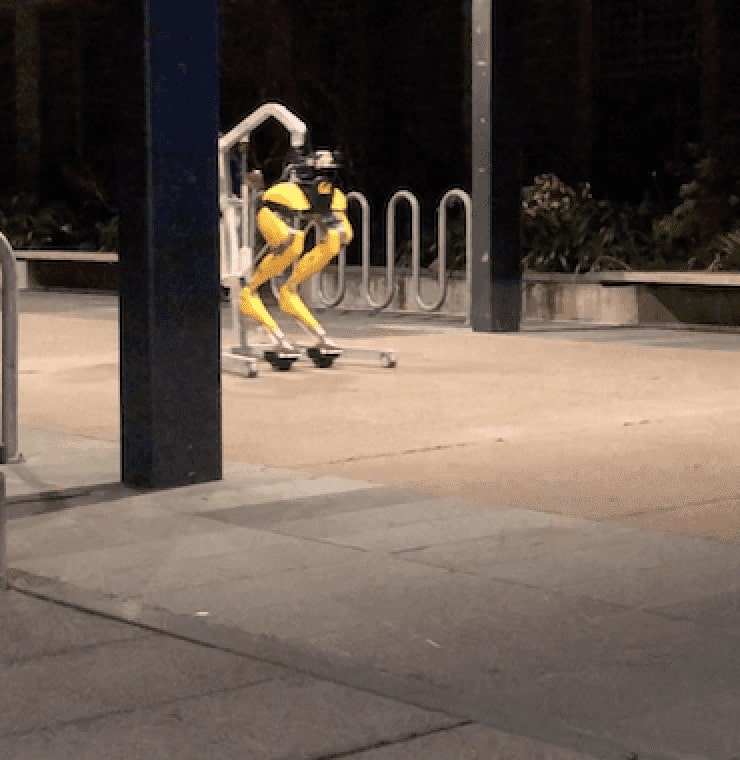}
    \end{subfigure}
    \begin{subfigure}[h]{0.32\linewidth}
        \centering
        \includegraphics[width=1.0\linewidth]{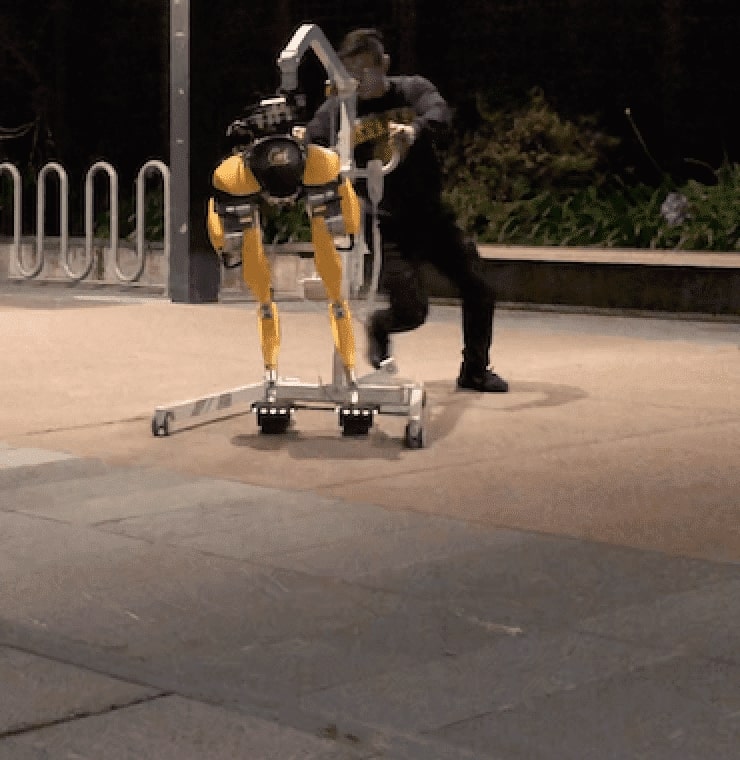}
    \end{subfigure}
    \begin{subfigure}[h]{0.32\linewidth}
        \centering
        \includegraphics[width=1.0\linewidth]{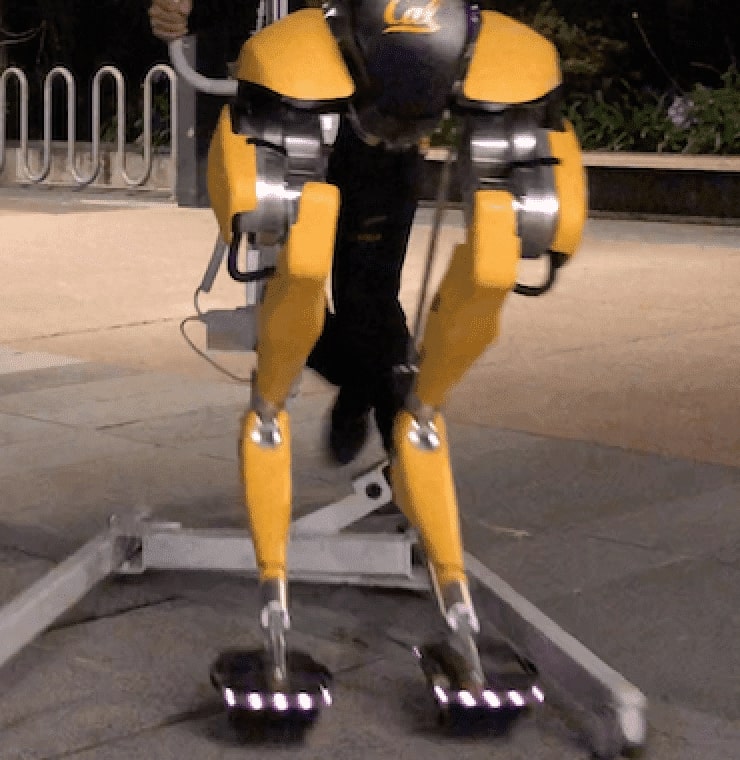}
    \end{subfigure} 
    \caption{Cassie executing a fast turn by leaning, with the lean angle directly dependent on the forward and rotational velocities.}
    \label{turning}
\end{figure}

A similar step input for turning yaw rotational velocity of \SI{0.5}{rad/s} and \SI{0.0}{rad/s} produced the response shown in Fig.~\ref{cassie_vel}.  Clearly, the tracking is not perfect and has a steady-state tracking error with RMSE is about \SI{0.1446}{rad/s}. This can be improved with better choice of gains or incorporating an integral feedback.
Furthermore, Fig.~\ref{turning} shows Cassie performing a high-speed turn. It is clear from the center graphic of Fig. \ref{turning} that Cassie successfully leans into the turn by shifting its COM position along the y-axis in order to prevent from tipping over during the turn.


\subsection{Wave Pattern with Feet for Obstacle Avoidance}
\begin{figure}
    \centering
    \begin{subfigure}[t]{0.32\columnwidth}
        \centering
        \includegraphics[width=\textwidth]{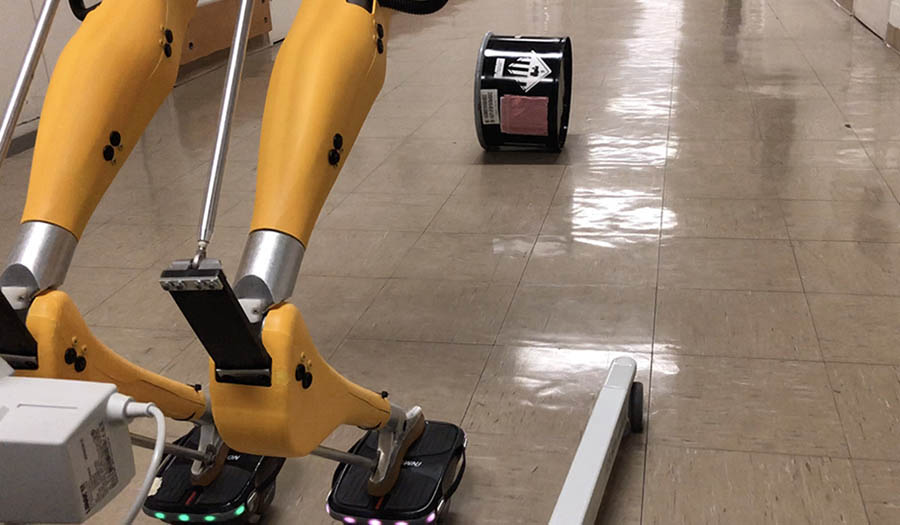}
        \label{featherstone_vel}
    \end{subfigure}
    \begin{subfigure}[t]{0.32\columnwidth}
        \centering
        \includegraphics[width=\textwidth]{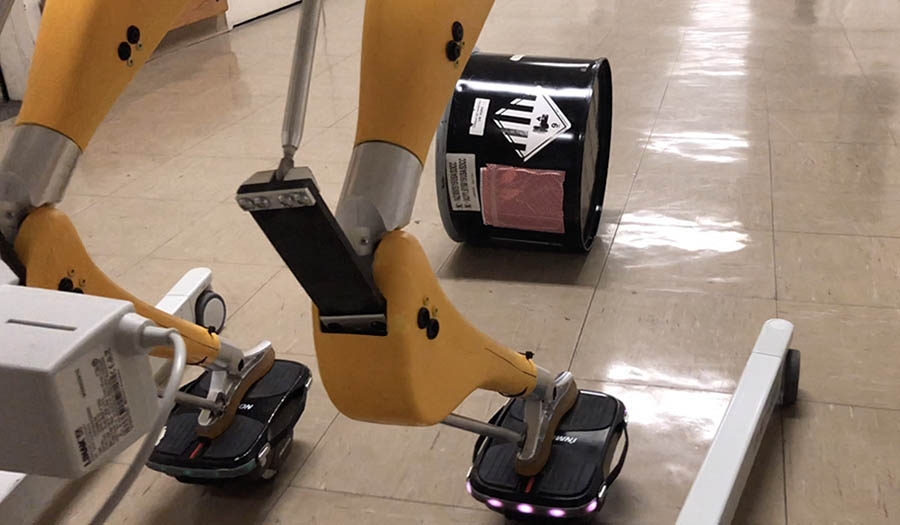}
        \label{sim_turn}
    \end{subfigure}   
    \begin{subfigure}[t]{0.32\columnwidth}
        \centering
        \includegraphics[width=\textwidth]{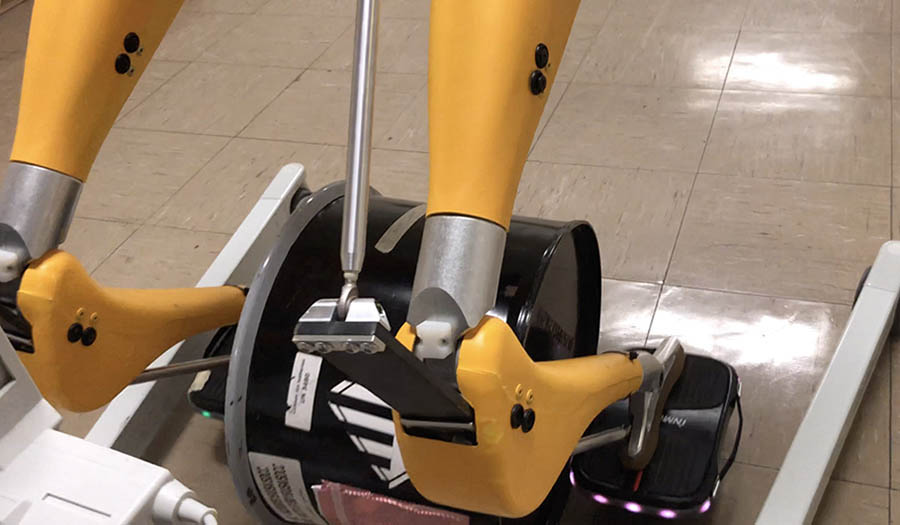}
        \label{none}
    \end{subfigure} 
    \caption{Cassie showing wave pattern with feet task by avoiding an obstacle without going around it. For certain obstacles, Cassie can stay on a straight trajectory while still missing the obstacle.} 
    \label{abduction_obs}
    \vspace{-5mm}
\end{figure}

Certain obstacles are short enough where Cassie does not need to circumnavigate them in order to avoid collision. With the Hovershoes, Cassie can split around the obstacle using the wave pattern controller. 
Fig.~\ref{abduction_obs} shows that Cassie successfully avoids an obstacle by doing a wave pattern maneuver with its feet. The decision making of splitting its feet was made by a human operator.

\subsection{Obstacle Avoidance}
\begin{figure}
    \centering
    \begin{subfigure}[h]{0.32\linewidth}
        \centering
        \includegraphics[width=1.0\linewidth]{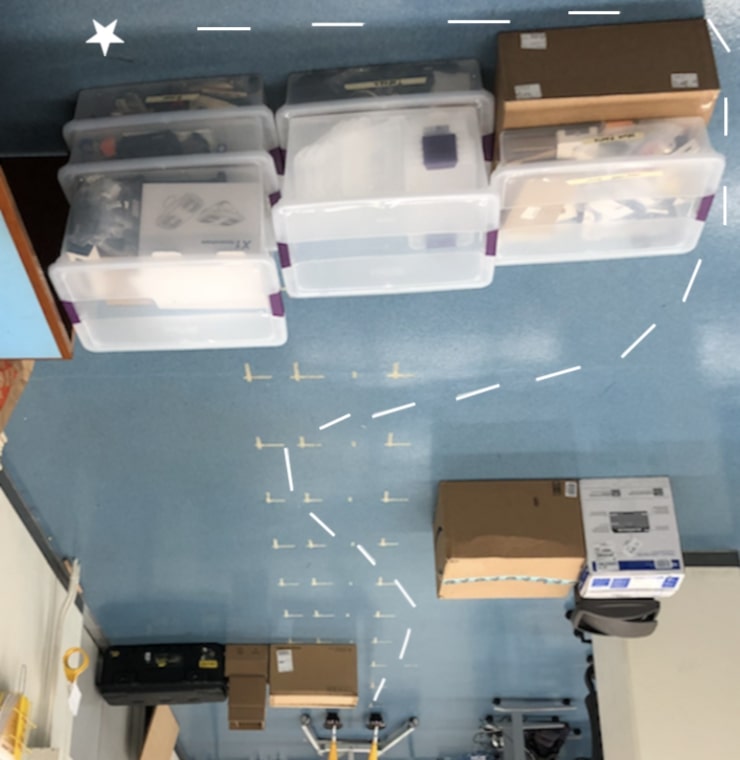}
    \end{subfigure} 
    \begin{subfigure}[h]{0.32\linewidth}
        \centering
        \includegraphics[width=1.0\linewidth]{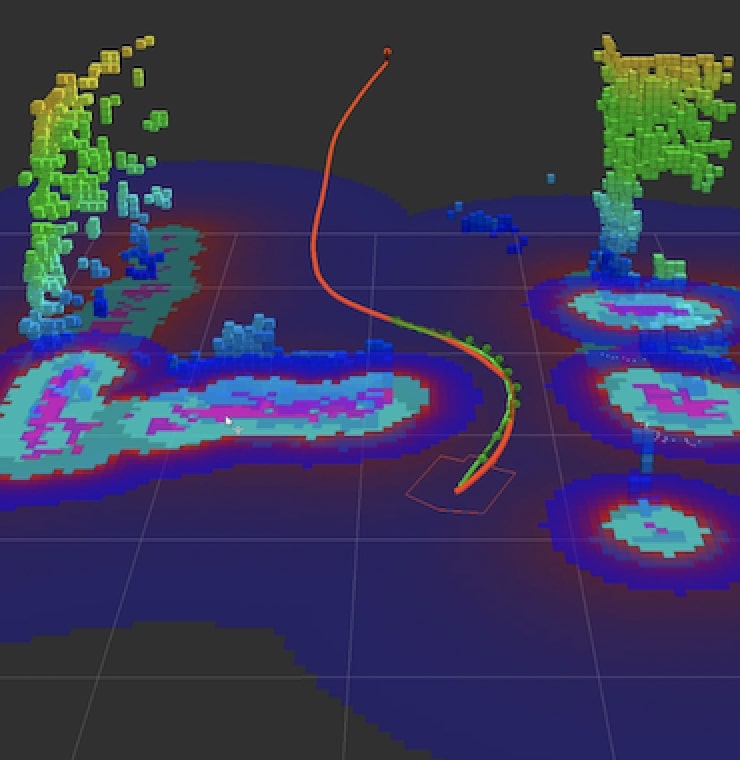}
    \end{subfigure} 
    \begin{subfigure}[h]{0.32\linewidth}
        \centering
        \includegraphics[width=1.0\linewidth]{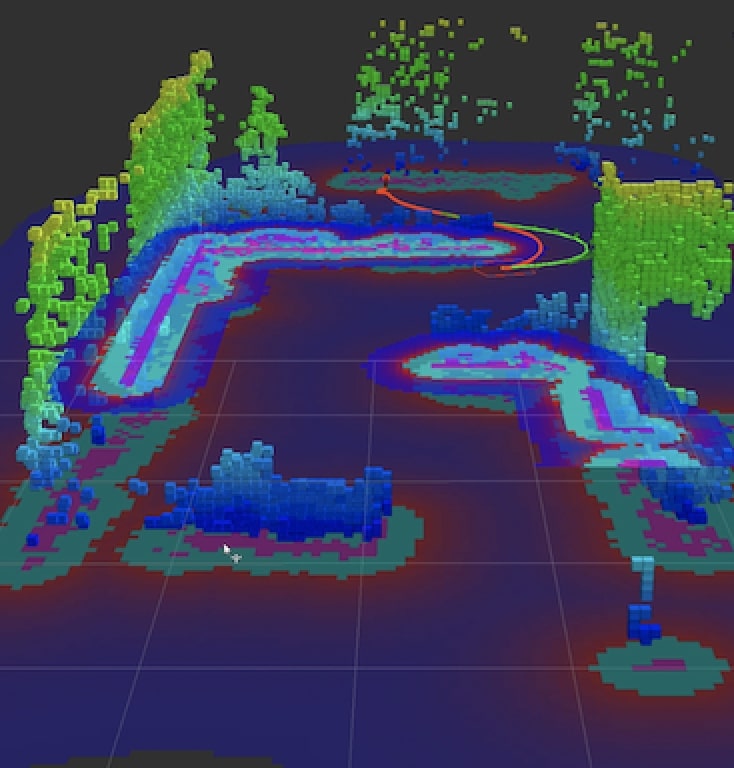}
    \end{subfigure} 
    \caption{Cassie performing obstacle avoidance. The leftmost figure is a picture of the obstacle course. The rest of the pictures are sequential snapshots of the path planner and 3-D map while the robot is autonomously navigating through the course.
    }
    \label{obstacle}
    \vspace{-4mm}
\end{figure}

Finally, combining our controllers with the computer vision obstacle detection system and path planner, we achieved autonomous locomotion through an obstacle course. The leftmost graphic of Fig.~\ref{obstacle} shows the obstacle course and the rest show Cassie autonomously navigating through the obstacle course.

\section{Shortcomings of our Proposed Work}
\label{sec:Limitations}

Our proposed method is the simplest control and autonomy package for achieving the task of riding Hovershoes.  In particular, our controller decouples the system dynamics into several simple sub-components, our planner only considers a kinematical model of the system, and our vision system only uses the stereo camera and IMU without using the additional sensors on the robot.  By considering a fully coupled dynamical system and incorporating the dynamics into the planner and taking all available sensing into account, we can further improve our state estimation, controller and planner performance significantly.  Finally, our proposed solution assumes Cassie is somehow initialized on the Hovershoes, which is currently done manually and does not truly offer multi-modal locomotion capabilities yet.

\section{Conclusions}
\label{sec:Conclusion}
In this paper, we have presented a framework for autonomous locomotion of a Cassie bipedal robot over Hovershoes.  
Our developed framework enables the Cassie bipedal robot to interact with the Hovershoes to balance, regulate forward and rotational velocity, achieve fast turns, and move over flat terrain, slopes, stairs, and rough outdoor terrain, as well as autonomously navigate an obstacle course.

The future work will focus on combining Cassie walking to achieve multi-modal locomotion. Currently human operators are needed to initialize Cassie on the Hovershoes. We would like to develop a method for Cassie to step up on the Hovershoes without the need of operators. 

\balance


\addtolength{\textheight}{-12cm}   



\balance

\bibliographystyle{IEEEtranS}
\bibliography{ref}

\end{document}